\documentclass[lettersize,journal]{IEEEtran}
\usepackage{amsmath,amsfonts}
\usepackage{amsthm}
\usepackage{algorithm}
\usepackage{array}
\usepackage[caption=false,font=footnotesize,labelfont=rm,textfont=rm]{subfig}
\usepackage{textcomp}
\usepackage{stfloats}
\usepackage{url}
\usepackage{verbatim}
\usepackage{graphicx}
\usepackage{cite}
\usepackage{color, soul} 
\usepackage{booktabs}
\usepackage{multirow}
\usepackage[pagebackref=false,breaklinks=false,linkcolor=red,anchorcolor=black, citecolor=green,colorlinks,bookmarks=true]{hyperref}
\setulcolor{blue} 

\usepackage{mdwmath}
\usepackage{mdwtab}
\usepackage{eqparbox}
\usepackage{algpseudocode}
\usepackage{bm}
\usepackage{makecell}
\usepackage{pdflscape}
\usepackage{afterpage}
\usepackage{bbding}
\usepackage{lscape}
\usepackage{booktabs}
\usepackage{tikz}
\usepackage{bbding}
\usepackage{pgfplots}
\usepackage{caption}
\usepackage{rotating}
\usepackage{tabularray}
\usepackage{tabularx}
\usepackage{ragged2e}
\usepackage{threeparttable}
\newcolumntype{L}{>{\RaggedRight\hangafter=1\hangindent=0em}X}
\newcolumntype{C}{>{\Centering\hangafter=1\hangindent=0em}X}

\usepackage{pgfplots}
\pgfplotsset{compat=1.7}

\begin{document}




\title{\textcolor{black}{SARATR-X}: Toward Building A Foundation Model for SAR Target Recognition}
\author{Weijie Li, Wei Yang$^{\ast}$, Yuenan Hou, Li Liu$^{\ast}$, Yongxiang Liu$^{\ast}$, Xiang Li 
\thanks{
This work was supported by the National Natural Science Foundation of China under Grant 62376283, the Science Fund for Distinguished Young Scholars of Hunan Province under Grant 2024JJ2066, the Science and Technology Innovation Program of Hunan Province under Grant 2022RC1092, and the Key Stone Grant JS2023-03 of the National University of Defense Technology.
\emph{($^{\ast}$Corresponding authors: Wei Yang, Li Liu, and Yongxiang Liu.)}}
\thanks{Weijie Li, Wei Yang, Li Liu, Yongxiang Liu, and Xiang Li are with the College of Electronic Science and Technology, National University of Defense Technology, Changsha, 410073, China (e-mail:  lwj2150508321@sina.com, \textcolor{black}{yw850716@sina.com, liuli\_nudt@nudt.edu.cn, lyx\_bible@sina.com, and lixiang01@vip.sina.com.}).}
\thanks{Yuenan Hou is with the Shanghai AI Laboratory, Shanghai, 200000, China \textcolor{black}{(e-mail: Yuenan\_Hou@163.com)}.}
}
\markboth{In preparation for submission}%
{Li \MakeLowercase{\textit{et al.}}: SARATR-X}

\maketitle

\begin{abstract}
Despite the remarkable progress in synthetic aperture radar automatic target recognition (SAR ATR), recent efforts have concentrated on \textcolor{black}{detecting and classifying} a specific category, \emph{e.g.}, vehicles, ships, airplanes, or buildings. One of the fundamental limitations of the top-performing SAR ATR methods is that the learning paradigm is supervised, task-specific, limited-category, closed-world learning, which depends on massive amounts of accurately annotated samples that are expensively labeled by expert SAR analysts and have limited generalization capability and scalability. In this work, we make the first attempt towards building a foundation model for SAR ATR, termed \textbf{SARATR-X}. \textbf{SARATR-X} learns generalizable representations via self-supervised learning (SSL) and provides a \textcolor{black}{cornerstone} for label-efficient model adaptation to generic SAR target detection and classification tasks. Specifically, SARATR-X is trained on 0.18 M unlabelled SAR target samples, which are curated by combining contemporary benchmarks and constitute the largest publicly available dataset till now. Considering the characteristics of SAR images, a backbone tailored for SAR ATR is carefully designed, and a two-step SSL method endowed with multi-scale gradient features was applied to ensure the feature diversity and model scalability of SARATR-X. The capabilities of SARATR-X are evaluated on classification under few-shot and robustness settings and detection across various categories and scenes, and impressive performance is achieved, often competitive with or even superior to prior fully supervised, semi-supervised, or self-supervised algorithms. Our SARATR-X and the curated dataset are released at \url{https://github.com/waterdisappear/SARATR-X} to foster research into foundation models for SAR image interpretation. 
\end{abstract}


\section{Introduction}
\label{Introduction}
\IEEEPARstart{S}{ynthetic} aperture radar (SAR)~\cite{sun2021spaceborne,gagliardi2023satellite,ref1, tsokas2022sar}, developed from electromagnetic scattering in microwave bands, plays a crucial role in active Earth observations, functioning effectively across diverse weather conditions and lighting environments. With the rapid development of SAR imaging techniques, high-resolution SAR images can be accessed more easily than before, enabling even more research opportunities for intelligent interpretation of SAR images. SAR automatic target recognition (ATR), aiming at automatically localizing and classifying objects of interest (\emph{e.g.}, vehicles, ships, airplanes, or buildings) in SAR images, is a longstanding, important yet challenging problem in SAR image intelligent interpretation~\cite{zhu2021deep, huang2022physically,wang2022self, datcu2023explainable}. SAR ATR\footnote{\textcolor{black}{In this paper, we focus on SAR ATR's classification and detection tasks. The classification aims to recognize the class and type of a single object, and the detection identifies the class and location of a single or multiple objects.}} plays an essential role in civil and national defense applications such as modern airport management, disaster management, urban planning and infrastructure monitoring, military reconnaissance, and maritime surveillance. Therefore, it has become an active research area for several decades~\cite{wang1998recognition,o2001sar,sun2007adaptive, ref7,9915465,10050159}. In the past decade, deep learning has brought tremendous success for SAR ATR~\cite{li2022deep, kechagias2021automatic, li2023comprehensive}. \textcolor{black}{Despite the significant progress, the following fundamental challenges will need to be addressed to advance the field of SAR ATR.}

\begin{figure}[!tb]
\centering
\includegraphics[width=8.8cm]{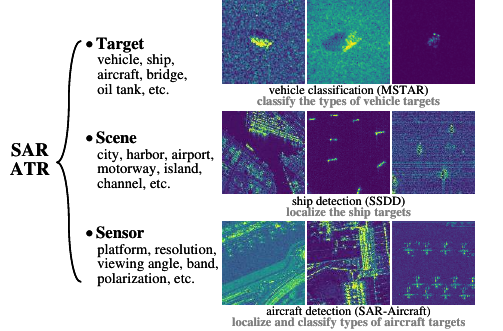}
\caption{\textcolor{black}{Various specialized SAR ATR datasets and tasks}. SAR ATR includes various imaging conditions \textcolor{black}{(\emph{i.e.} operating condition), such as} targets, scenes, and sensors. However, the datasets are often collected in specific settings for certain tasks due to high costs. For example, MSTAR~\cite{MSTAR} is a ten-type vehicle target classification dataset in the X-band and grass scenarios, and SAR-Aircraft is a seven-type aircraft detection dataset collected from three airports and a C-band satellite. Specialized algorithms have been proposed for these datasets. However, the differing target characteristics, scene information, and sensor parameters have complicated the generalization of existing algorithms. As such, this paper aims to develop \emph{a SAR ATR foundation model}, a generalized method for conducting various tasks.}
\label{fig_problem}
\end{figure}

Firstly, \textbf{task-specific property.} One of the fundamental limitations of the current ATR methods~\cite{10283916,feng2022electromagnetic,fu2021scattering, fu2021anchor,sun2022scan} is that one model is trained and evaluated on one specific task. The detection and classification of a specific coarse category (\emph{e.g.}, vehicles, ships, airplanes, or buildings in Fig.~\ref{fig_problem}.) all require their own deep models. As a result, the task-specific properties of these deep models pose significant challenges for training new tasks or developing a comprehensive SAR ATR system since each task must be learned independently from the ground up, requiring vast amounts of labeled data. This results in computational inefficiency, lower accuracy, and inconsistent results between the different models. Secondly, \textbf{heavy reliance on supervised learning.} Recent progress in SAR ATR~\cite{zhang2022sefepnet,zhou2024diffdet4sar,10753051}, while substantial, has been limited to supervised learning, which heavily depends on massive amounts of accurately annotated target samples that are expensively labeled by expert SAR analysts and have limited generalization capability and scalability. However, the scarcity of expert SAR analysts cannot meet such an exhaustive requirement,
leaving vast amounts of SAR images unlabelled and unexploited. 
Thirdly, \textbf{the ignorance of SAR image characteristics in model designs.} The imaging characteristics of SAR imagery differ significantly from those of optical imagery, leading to a significant domain gap between natural and SAR images. This raises significant challenges when one intends to transfer prior knowledge from the natural image domain. Different strong prior knowledge of SAR imagery, including speckle noise, discrete target appearances, and the lack of geometry, texture, and contour cues, needs special consideration when designing backbone architectures and \textcolor{black}{learning strategies}. Most of the current mainstream backbones and methods designed on natural images are not suitable for the aforementioned information.
Finally, \textbf{underdeveloped open-source ecosystem.} Due to data sensitivity, the open-source ecosystem across the entire field is underdeveloped, making it challenging to share code and data publicly. Currently, there are no large and representative benchmark datasets for SAR ATR. As a result, this locks the potential of recent deep learning techniques for SAR ATR and significantly slows down the development of this field.


\IEEEpubidadjcol

Recently, the remarkable success of foundation models (FMs)~\cite{ChatGPT, radford2021learning,kirillov2023segment,bai2023sequential} has led to a learning paradigm shift in artificial intelligence. Foundation models~\cite{bommasani2021opportunities}, pretrained on extensive data in a task-agnostic manner (generally via self-supervised learning), can be flexibly adapted to a wide range of downstream tasks. Self-supervised learning (SSL)~\cite{liu2021self, balestriero2023cookbook,jing2020self,10533864,xiao2024highly} can be used to mitigate label inefficiencies by exploring supervision in the data directly, thereby reducing the reliance on expensive expert labeling while efficiently scaling the data and models. FMs shine in a broad range of areas, including natural language processing, computer vision, speech recognition, and medical image analysis. 
\textcolor{black}{As summarized in Table~\ref{table_relatexwork}}, FMs have also been explored in remote sensing image understanding, but \textcolor{black}{they are} mostly limited to the evaluation of optical data. To our knowledge, the huge potential of FMs for SAR image interpretation remains completely locked. 

In our preliminary work, a novel SSL method for SAR imagery named SAR Joint-Embedding Predictive Architecture (SAR-JEPA)~\cite{li2023self} was proposed and demonstrated promising results. 
Given the aforementioned discussion, to fully unlock the potential of FMs for SAR image interpretation, we present the first attempt toward building a foundation model for SAR ATR and propose \textbf{SARATR-X} that learns generalizable representations via SSL and provides a basis for label-efficient model adaptation to generic SAR target detection and classification tasks. While conceptually simple, existing methods for building FMs in other domains cannot be just directly applied to SAR ATR. We are still facing significant challenges, which are discussed and addressed below.

\textbf{Pre-training datasets} must include diverse target categories and imaging conditions to accommodate various downstream tasks. However, SAR ATR lacks a large-scale dataset, such as ImageNet~\cite{fei2022searching}, and the most \textcolor{black}{used} MSTAR dataset only includes fine-grained vehicle categories that are not suitable for larger-scale pre-training (see Table~\ref{table_relatexwork2}). 
As such, SARDet-100K~\cite{li2024sardet100k} incorporates 9 SAR target detection datasets. With the number of open-source SAR ATR datasets increasing, most of the datasets were integrated into this study as part of the pre-training. A total of 14 classification and detection datasets with different target categories and imaging conditions were included as a new pre-training dataset, SARDet-180K, to explore the potential of FMs, as seen in Table~\ref{table_dataset}.

\textbf{Model backbones} aim to achieve better spatial resolution representations in remote sensing images, especially for small targets in large imagery. Transformers and convolutional neural networks are among the most common architectures for these tasks.
\textcolor{black}{As shown in Table~\ref{table_relatexwork}, the Transformer offers better spatial resolution without downsampling, which is most commonly used in recent studies.} As such, HiViT~\cite{zhang2023hivit} was selected as it has the advantages of a Swin transformer high-resolution input and can drop patches in masked image modeling (MIM).

\textbf{Self-supervised learning} is complicated by SAR image quality, which is negatively affected by speckle noise in coherent imaging. The coherent imaging resulting visual features are also not as distinct or rich as in nature RGB images. Contrastive learning~\cite{zhai2022weakly,pei2023self} uses data augmentation and preprocessing to reduce noise, while MIM~\cite{wang2023feature,li2023self,li2024sardet100k} applies various target feature for guided signals to suppress noise (see Table~\ref{table_relatexwork2}). 
As such, the primary task of SAR SSL is to enhance the quality of feature learning and guide signals. For example, PGIL~\cite{huang2022physically} leveraged a sub-frequency feature of complex SAR images to learn physics information, while our SAR-JEPA~\cite{li2023self} applied multi-scale gradient ratios to solve for multiplicative speckle noise and capture target shapes. Furthermore, multi-stage training~\cite{li2024sardet100k} from ImageNet to SAR diminished the effects of noise on model diversity, as seen in Fig.~\ref{visual_attention_distance}. Thus, we applied two-step pre-training from ImageNet to SAR to increase model diversity during pre-training with SAR images. Besides, multi-scale gradient features were used as high-quality guide signals for MIM with SAR images.

\textbf{Evaluation tasks} need to comprehensively evaluate the performance of a foundation model for different tasks and settings. Three open-source target datasets were utilized by first constructing a fine-grained classification dataset, SAR-VSA, with 25 categories to evaluate the effectiveness of the proposed improvements. A comprehensive comparison was then performed between the proposed SARATR-X and existing methods for public classification and detection tasks. 

SARATR-X achieved superior performance in 5 datasets across 8 task settings, as shown in Fig.~\ref{fig_sota}, which is competitive with prior methods on various SAR ATR tasks (\textcolor{black}{classification with few-shot and robustness setting and} detection with specific categories or various categories). We hope that this work could advance the development of the intersection of general SAR target recognition and foundation models. 

The primary contributions of this study can be summarized as follows:

\begin{figure}[!tb]
\centering
\includegraphics[width=8.8cm]{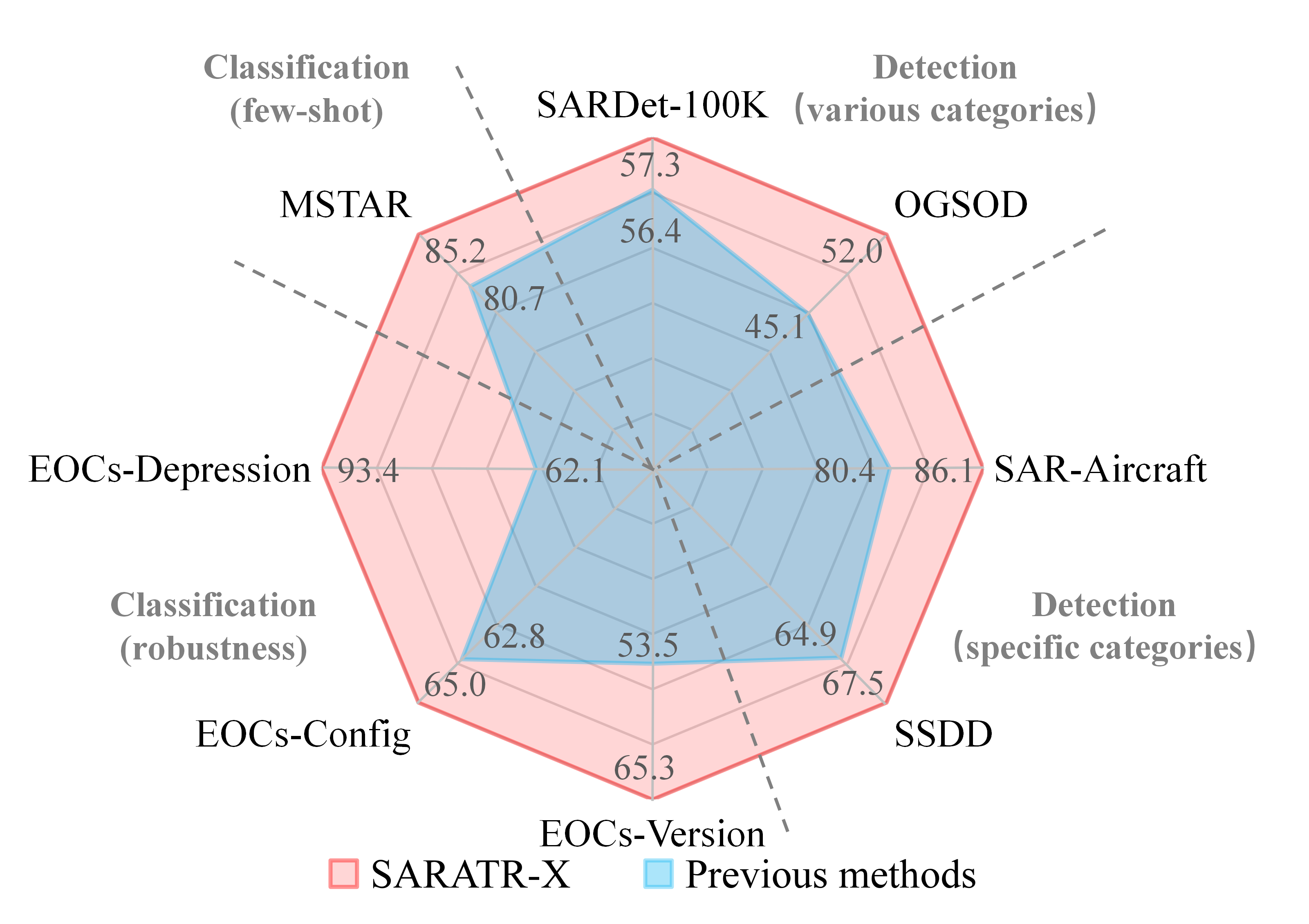}
\caption{\textcolor{black}{Results on classification and detection tasks.} SARATR-X performed well across 5 datasets with 8 settings. It was superior to existing SSL methods (BIDFC~\cite{zhai2022weakly}) for target classification in the fine-grained vehicle MSTAR dataset~\cite{MSTAR} with a few-shot setting. In addition, it performed well under extended operating conditions (EOCs)~\cite{zhang2021domain} (\emph{i.e.}, imaging conditions with variable depression angle (EOCs-Depression), target configuration (EOCs-Config), and version (EOCs-Version)). SARATR-X also demonstrated competitive object detection performance with existing supervised methods applied to various categories (SARDet-100K~\cite{li2024sardet100k} and OGSOD~\cite{wang2023category}), as well as specific categories for ships (SSDD~\cite{zhang2021sar}) and aircraft (SAR-AIRcraft~\cite{wang2023sar}). Our study shows the potential of a foundation model for SAR ATR.}
\label{fig_sota}
\end{figure}

\begin{enumerate}
\item[$\bullet$] 
We present the first foundation model called SARATR-X, which learns generalizable representations via SSL from large-scale unlabelled data and provides a cornerstone for generic SAR target detection and classification tasks.
\item[$\bullet$] 
We systematically investigate a foundation model framework for SAR ATR. We build the largest publicly available pre-training dataset, SARdet-180k, and fully discuss the model architecture and proposed SSL method with many comparisons.
\item[$\bullet$] 
SARATR-X is evaluated comprehensively with various SAR ATR tasks, such as few-shot classification, robust classification, ship detection, aircraft detection, and detection with various categories. 
\end{enumerate} 

The remainder of this paper is organized as follows. Sec.~\ref{Related Work} introduces related work in remote sensing and SAR ATR. Sec.~\ref{Approach} discusses the proposed foundation model (SARATR-X). Secs.~\ref{Experiments} and~\ref{Leveraging} conduct extensive experiments to demonstrate the superiority of the proposed method. Sec.~\ref{Conclusion} concludes the paper and discusses future work.

\section{Related Work}
\label{Related Work}

\begin{table*}[tb]
\centering
\caption{\textcolor{black}{A list of visual FMs used for remote sensing. There are many FMs are available for various modalities and tasks, but existing research (FG-MAE, SkySense, and OFA-Net) has focused on scene-level tasks with the SAR modality. SARATR-X represents an extension of our previous work (SAR-JEPA and MSFA~\cite{li2023self,li2024sardet100k}) and advances the state of FMs for SAR ATR. SSL methods can be divided into two categories: contrastive (i.e., contrastive learning used to obtain invariant features) and generative (i.e., masked image modeling used to generate and predict features). SD: stable diffusion. Swin: Swin transformer.}}
\label{table_relatexwork}
\renewcommand\arraystretch{1.3}
\resizebox{0.95\linewidth}{!}{%
\begin{tabular}{lcp{1.7cm}<{\Centering} p{2.4cm}<{\Centering} p{1.3cm}<{\Centering} p{1.7cm}<{\Centering} p{6.5cm}<{\RaggedRight}} 
\toprule
\multicolumn{1}{c}{\multirow{1}{*}{\textbf{Foundation Model}}} & \multirow{1}{*}{\textbf{Year}} & \multirow{1}{*}{\textbf{Modality}} & \textbf{Dataset} & \textbf{Backbone} & \textbf{SSL} & \multicolumn{1}{c}{\multirow{1}{*}{\textbf{Tasks}}} \\ 
\cmidrule(lr){1-7}
SatMAE~\cite{cong2022satmae} & 2022 & Multi-spectral, RGB & fMoW RGB/Sentinel & ViT & Generative & Scene classification and semantic segmentation\\
RVSA~\cite{wang2022advancing} & 2022 & RGB & MillionAID & ViT, ViTAE & Generative & Scene classification, object detection, and semantic segmentation\\
RingMo~\cite{sun2022ringmo} & 2022 & RGB & Self-built & ViT, Swin & Generative & Scene classification, object detection, semantic segmentation, and change detectionn\\
RingMo-Sense~\cite{yao2023ringmo} & 2023 & RGB & Self-built & Video Swin & Generative & object detection and tracking, images segmentation, and 3 prediction tasks\\
CMID~\cite{muhtar2023cmid} & 2023 & RGB & MillionAID & ResNet, Swin & Generative \& Contrastive & Scene classification, object detection, and semantic segmentation\\
GFM~\cite{mendieta2023towards} & 2023 & RGB & GeoPile & Swin & Generative & Scene classification, semantic segmentation, change detection, and super-resolution\\
DiffusionSat~\cite{khanna2023diffusionsat} & 2023 & RGB & fMoW, Satlas, SpaceNet & SD & Generative & Image generation, super-resolution, temporal generation, and in-painting\\
Scale-MAE~\cite{10377166} & 2023 & RGB & fMoW RBG & ViT & Generative & Scene classification, object detection, and semantic segmentation\\
FG-MAE~\cite{wang2023feature} & 2023 & Multi-spectral, SAR & SSL4EO-S12 & ViT & Generative & Scene classification and semantic segmentation\\
SMLFR~\cite{dong2024generative} & 2024 & RGB & GeoSense & ConvNeXt & Generative & Object detection, semantic segmentation, and change detection\\
SkySense~\cite{guo2023skysense} & 2024 & 3 modalities & HSROIs, TMsI, TSARI & Swin, ViT & Contrastive & Scene classification, object detection, semantic segmentation, and change detection\\
OFA-Net~\cite{xiong2024one} & 2024 & 4 modalities & 5 multi-modalities datasets & ViT & Generative & Scene Classification and semantic segmentation\\
\cmidrule(lr){1-7}
SARATR-X & 2024 & SAR & ImageNet \& 14 SAR datasets & HiViT & Generative & Target classification and object detection\\
\bottomrule
\end{tabular}}
\end{table*}

\begin{table*}[tb]
\centering
\caption{\textcolor{black}{Related SSL work in SAR ATR. They has involved the use of various datasets, architectures, and SSL algorithms. Inspired by previous work, this study presents the first SAR ATR foundation model, SARATR-X.}}
\label{table_relatexwork2}
\renewcommand\arraystretch{1.3}
\resizebox{0.95\linewidth}{!}{%
\begin{tabular}{lcp{2cm}<{\Centering} p{2.0cm}<{\Centering} c p{4.0cm}<{\RaggedRight} p{4.8cm}<{\RaggedRight}} 
\toprule
\multicolumn{1}{c}{\multirow{1}{*}{\textbf{Method}}} & \multirow{1}{*}{\textbf{Year}} & \textbf{Dataset} & \textbf{Backbone} & \textbf{SSL} & \multicolumn{1}{c}{\multirow{1}{*}{\textbf{Tasks}}} & \multicolumn{1}{c}{\multirow{1}{*}{\textbf{Description}}}\\ 
\cmidrule(lr){1-7}
RotANet~\cite{wen2021rotation} & 2021 & MSTAR & CNN & Generative & Target classsification (MSTAR) & Predicting rotational patterns of targets as a loss regularization term for classification task.\\
UACL~\cite{xu2021adversarial} & 2021 & MSTAR, FUSAR-Ship & CNN & Contrastive & Target classsification (MSTAR) & Contrastive defense method to enhance model robustness to different adversarial attack.\\
PGIL~\cite{huang2022physically} & 2022 & Sea-ice, Urban & ResNet-18 & Contrastive & Scene classsification (sea-ice, urban) & Contrastive learning between sub-frequency features of complex images and deep features of amplitude images.\\
BIDFC~\cite{zhai2022weakly} & 2022 & MSTAR, OpenSARShip & ResNet-18 & Contrastive & Target classsification (MSTAR, OpenSARShip) & Weakly contrastive learning for pre-training in fine-grained datasets with similar sample.\\
TSCL~\cite{pei2023self} & 2023 & MSTAR & ResNet-50 & Contrastive & Target classsification (MSTAR, OpenSARShip) & Pre-processing of SAR images for noise removal to improve image quality for Contrastive learning.\\
FG-MAE~\cite{wang2023feature} & 2023 & SSL4EO-S12 & ViT-S & Generative & Scene classification (EuroSAT, BigEarthNet-MM), semantic segmentation (DFC2020) & Hand-craft feature HOG as the target signal for MIM in SAR images.\\
SAR-JEPA~\cite{li2023self} & 2024 & MSAR, SAR-Ship, SARSim, SAMPLE & ViT-B & Generative & Target classification (MSTAR, FUSARShip, SAR-ACD) & Local reconstruction and multi-scale gradient feature for MIM with small object in noise SAR images.\\
MSFA~\cite{li2024sardet100k} & 2024 & ImageNet, DOTA, SARDet-100k & ConvNeXt, Swin-B, Van & Generative & Object detection (SARDet-100K, SSDD, HRSID) & Multi-stage pre-training strategy for MIM from RGB images to SAR images.\\
SARATR-X & 2024 & ImageNet, 14 SAR target datasets & HiViT-B & Generative & Target classification and object detection (5 datasets and 8 settings) & A first foundation model for target recognition in SAR Images.\\
\bottomrule
\end{tabular}}
\end{table*}

Visual foundation models are actively being used for remote sensing applications, and different algorithms have recently been proposed for various modalities and tasks. This study focuses on a foundation model used for SAR ATR (\emph{i.e.}, SAR image-based target classification and object detection). In the following sections, we introduce recent developments in remote sensing and \textcolor{black}{SAR} foundation models. 

\subsection{Foundation models for remote sensing}
Remote sensing foundation models~\cite{10254282,lu2024ai} have received widespread attention in recent years and have achieved effective learning across various modalities and tasks. Many of these studies have used existing large-scale pre-training datasets or have collected large quantities of samples from different sources. Model backbones have been established by improving attention mechanisms, positional encoding, and other aspects used to enhance the perception of complex spatial information. MIM has also been used to learn spatial-temporal contextual information, while contrast learning has been applied to multi-modal learning.

SatMAE~\cite{cong2022satmae} involves a novel masking strategy with the temporal and spectral positional encoding used for multi-spectral and temporal images. SatMAE has also achieved excellent performance in scene classification and semantic segmentation tasks using a new dataset (fMoW Sentinel) that includes 13 different frequency bands. 
RVSA~\cite{wang2022advancing} improves a pre-training ViT backbone using a rotated variable-size window attention method for arbitrarily oriented objects. This work demonstrates the importance of learning complex spatial contextual relationships for targets in remote sensing images.
RingMo~\cite{sun2022ringmo} utilizes a patch incomplete mask strategy for dense or small objects and a self-constructed set of 2 million images, proving effective in a variety of tasks. It shows the potential of large-scale pre-training.
RingMo-Sense~\cite{yao2023ringmo} offers a three-branch network and a masking strategy for modeling spatio-temporal interactions in temporal images. 
CMID~\cite{muhtar2023cmid} combines contrastive learning and masked image modeling to learn global semantic information and local spatial information. This work also shows the importance of learning the diversity of contextual relationships in remote-sensing images.
GFM~\cite{mendieta2023towards} focuses on the differences between natural and remote sensing images and employs a multi-object continual pre-training approach to leverage information from both. It shows that detailed information in natural images can complement remote sensing images well.
DiffusionSat~\cite{khanna2023diffusionsat} is the first remote sensing generative model to employ geographic information embedding in stable diffusion. 
Scale-MAE~\cite{10377166} reconstructs images at different frequencies with improved positional encoding for ViT.
FG-MAE~\cite{wang2023feature} employs various hand-designed features to replace original pixels in the MIM, thereby improving the feature quality. Similarly, SMLFR~\cite{dong2024generative} uses a low-pass filter to eliminate high-frequency information from the image pixels. These studies show the necessity of SSL high-quality guide signals. As such, our work focuses on the issues of SAR image quality while considering the range and spatial resolution of remote sensing. 

A variety of multi-modal remote sensing FMs have also been developed, including SkySense~\cite{guo2023skysense} and OFA-Net~\cite{xiong2024one}. SkySense~\cite{guo2023skysense} adopts a multi-granularity contrastive learning method to learn representations for different modalities. A GEO-context prototype has also been applied to embed geographical contextual information. OFA-Net~\cite{xiong2024one} employs a shared transformer backbone for multiple modalities. In addition, vision-language models~\cite{10506064}, such as EarthGPT~\cite{zhang2024earthgpt}, SkyEyeGPT~\cite{zhan2024skyeyegpt}, and LHRS-Bot~\cite{muhtar2024lhrs} incorporate large language models into various remote sensing image modalities. However, due to the difficulty of annotating SAR images, the collection of public datasets used by EarthGPT only contains 10,554 SAR ship images, much less than the 84,838 infrared or 907,945 optical images. As a result, this study explores visual FMs based on \textcolor{black}{unlabeled SAR images to improve the SAR ATR model's scalability with large-scale samples}.

\subsection{Related SSL in SAR}
SSL for SAR ATR has been investigated in multiple studies, \textcolor{black}{as detailed in Table~\ref{table_relatexwork2}.}
Early SSL was often used as a \textcolor{black}{auxiliary task} for classification tasks, as discussed below. 
RotANet~\cite{wen2021rotation} predicts the rotational patterns of MSTAR vehicle targets by capturing azimuthal features for classification tasks. 
UACL~\cite{xu2021adversarial} combines data augmentation and adversarial samples for contrastive learning to improve model robustness to various adversarial attacks. 
PGIL~\cite{huang2022physically} employs contrastive learning between complex SAR image sub-frequency features and deep amplitude image features, incorporating physical knowledge into classification tasks. SSL has also been used recently in model pre-training and fine-tuning frameworks. 
BIDFC~\cite{zhai2022weakly} proposes weakly contrastive learning for pre-training in fine-grained vehicle datasets (MSTAR) and applied Gaussian noise data augmentation to simulate SAR image noise. 
TSCL~\cite{pei2023self} applies SAR image pre-processing prior to data augmentation in contrastive learning. 
FG-MAE~\cite{wang2023feature} discusses different hand-crafted features for use with multi-spectral and SAR images and applies HOG features to SAR. In our previous studies, SAR-JEPA~\cite{li2023self} SAR-JEPA~\cite{li2023self} applies local reconstruction and multi-scale gradient features to collect target spatial signatures better. MSFA~\cite{li2024sardet100k} proposes a multi-stage process with a filter augmentation pre-training framework for use in large-scale RGB and SAR data detection. However, these studies have only explored classification or detection tasks on a small number of datasets. Inspired by these previous works, this study aims to systematically investigate the construction of a SAR ATR foundation model for various ATR datasets via SSL. 

\textbf{Our Insights -} These studies have demonstrated that SSL can achieve performance improvements across multiple categories~\cite{li2023self} and tasks~\cite{li2023self,li2024sardet100k}, which may be comparable to the performance of specially designed supervised methods~\cite{zhai2022weakly,li2024sardet100k}. However, it still lacks a foundation model for various SAR ATR applications. Besides, there is a lack of a pre-training and evaluation benchmark that contains images from classification and detection scenes. This inspired us to conduct systematic research into foundation models for general SAR target recognition, specifically with big data. We first extended the pre-dataset using different classification and detection tasks and scenarios (such as globally inland, marine, harbors, cities, and airports) based on existing research. A suitable model backbone is then discussed for the small target characteristics of remote sensing images. Since SSL requires high-quality guide signals from SAR images under the influence of noise, we applied two-step pre-training. Finally, we comprehensively evaluated the performance of the foundation models.

\section{Approach}
\label{Approach}
We aim to construct a foundation model for general ATR from large-scale SAR images via an SSL method. As described above, increasing SAR datasets and SSL studies have inspired us to develop a foundation model for SAR ATR. We focus on pre-training datasets, model backbones, SSL methods, and evaluation tasks to provide a systematic benchmark for SAR ATR foundation models.

\textcolor{black}{First, we establish a diverse dataset for pre-training and integrate 14 SAR target datasets to ensure that most SAR target samples are included and that samples are balanced across categories. Besides, based on target size and image resolution, we slice bigger image sizes than 1000 pixels to increase the number of samples and extract small target slices. Second, we compare different model architectures to select a suitable model architecture for remote sensing recognition. Third, we propose a two-step pre-training strategy to achieve diversity in attention distance for different target sizes and effective training and scaling in the presence of SAR image noise interference. Natural image pre-training weights are used as SAR image pre-training initialization to enhance attentional diversity and multi-scale gradient features (MGFs) to suppress speckle noise. Finally, we perform a linear probing evaluation of proposed SSL strategies on the self-constructed SAR classification dataset and compare SARATR-X with existing methods on the public SAR target recognition dataset.}

\subsection{Pre-training Dataset}

Previous research primarily employed MSTAR~\cite{MSTAR} as a pre-training dataset. While MSTAR provides high-quality vehicle targets, it only contains a few thousand commonly used samples. The images also suffer from background bias caused by a single imaging scene~\cite{li2023discovering}. In contrast, the ImageNet-1K pre-training set contains 1.4 million images with different categories and scenes. Since diverse target, scene, and sensor conditions constitute a large data sampling space in real-world scenarios, constructing a large pre-training dataset for the foundation model is central. 

The increasing availability of SAR target datasets is a primary motivation for achieving this goal. Although SAR images are expensive and no single dataset contains all popular target categories or imaging conditions, collecting target samples from various open-source datasets can still provide a pre-training set with distinct categories, scenes, and sensors. \textcolor{black}{Besides, considering that MSTAR's background bias is due to the strong correlation between target classes and acquisition locations, we need to integrate different target datasets to increase the diversity of scenes and sensor conditions while learning target contextual information.} As such, we constructed a new pre-training dataset, SARDet-180K, consisting of 186,600 SAR target samples from 14 open-source\footnote{In this paper, we did not consider target datasets that are not open source or datasets for SAR terrain classification. Furthermore, we did not choose all open-source ship dataset as pre-training to avoid too many marine scenarios~\textcolor{black}{causing class imbalance~\cite{rezvani2023broad}}, and other open-source ship datasets about 28,749 images were not added, including FUSAR-Ship~\cite{hou2020fusar}, LS-SSDD~\cite{zhang2020ls}, DSSDD~\cite{hu2021dual}, SRSDD~\cite{lei2021srsdd}, and RSDD~\cite{xu2022rsdd}. \textcolor{black}{Besides, the class imbalance can be alleviated using resampling and hard-sample mining during the pre-training and fine-tuning in practice. SARDet-180K includes 46,096 vehicles, 73,973 ships, 35,861 aircraft, and 46,830 other samples.}} SAR target datasets' all images, as described in Table~\ref{table_dataset}. This set aims, to the extent possible, to include common target categories (terrestrial and maritime targets such as vehicles, ships, aircraft, oil tanks, bridges, etc.), scenes (typical scenes such as cities, harbors, airports, oceans, etc.), and sensors (satellite, airborne, and simulation platforms of varying resolutions and bands). 

\begin{table*}[!tb]
\centering
\caption{\textcolor{black}{A description of the SARATR-X data, which included 14 open-source SAR datasets used for pre-training.} Large SAR imagery in the detection datasets contained more targets and scene types than the annotations. \textcolor{black}{Some large images in Sandia MiniSAR and SAR-AIRcraft were cropped to increase the number of available training samples based on their target size and image resolution. More detailed descriptions are in Appendix A.} Cls.: Classification. Det.: Detection. \# Img.: Number of images. \# Target: Number of target categories. \# Scene: Number of scenes. Res.: Resolution. Pol.: Polarization.}
\label{table_dataset}
\renewcommand\arraystretch{1.2}
\resizebox{\linewidth}{!}{%
\begin{tabular}{lcccccccccl} 
\toprule
\multicolumn{1}{c}{\textbf{Dataset}} & \textbf{Year} & \textbf{\textcolor{black}{Task}} & \textbf{\#~Imgs.} & \textbf{Img. Size} & \textbf{\# Targets} & \textbf{\# Scenes} & \textbf{\textcolor{black}{Res. (m)}} & \textbf{\textcolor{black}{Band}} & \textbf{\textcolor{black}{Pol.}} & \textbf{\textcolor{black}{Target description}} \\ 
\cmidrule(lr){1-11}
MSTAR~\cite{MSTAR} & 1995 & Cls. & 14,577 & $128\sim193$ & 10 & 1 & 0.3 & X & Single & Fine-grained vehicle dataset \\
Sandia MiniSAR~\cite{Sandia} & 2006 & Det. & 3,927 & 224 & $\geq1$ & $\geq7$ & 0.1 & Ku & Single & Terrestrial targets in urban, deserts, and others \\
SARSim~\cite{malmgren2017improving,kusk2016synthetic} & 2017 & Cls. & 21,168 & 139 & 14 & 3 & 0.3 & X & Single & Simulation vehicle dataset \\
SAMPLE~\cite{lewis2019sar} & 2019 & Cls. & 5,380 & 128 & 10 & 2 & 0.3 & X & Single & Simulation and measured vehicle dataset \\
SIVED~\cite{lin2023sived} & 2023 & Det. & 1,044 & 512 & $\geq1$ & $\geq4$ & $0.1\sim0.3$ & X/Ku/Ka & Single & Synthetic vehicle dataset \\
OpenSARShip~\cite{li2017opensarship} & 2017 & Cls. & 26,679 & $9\sim445$ & 14 & 10 & $2.3\sim17.4$ & C & Double & Fine-grained ship slices \\
SAR-Ship~\cite{ref54} & 2019 & Det. & 39,729 & 256 & $\geq1$ & $\geq4$ & $3\sim25$ & C & Quad & Ship dataset in complex scenes \\
AIR-SARShip~\cite{xian2019air} & 2019 & Det. & 801 & $512\sim1000$ & $\geq1$ & $\geq3$ & $1\sim3$ & C & Single & Ship dataset in complex scenes \\
HRSID~\cite{wei2020hrsid} & 2020 & Det. & 5,604 & 800 & $\geq1$ & $\geq2$ & $0.5\sim3$ & C/X & Quad & Instance-level ship dataset \\
SSDD~\cite{zhang2021sar} & 2021 & Det. & 1,160 & $214\sim668$ & $\geq1$ & $\geq2$ & $1\sim15$ & C/X & Quad & Ship dataset \\
SADD~\cite{zhang2022sefepnet} & 2022 & Det. & 883 & 224 & $\geq1$ & $\geq2$ & $0.5\sim3$ & X & Single & Aircraft dataset \\
SAR-AIRcraft~\cite{wang2023sar} & 2023 & Det. & 18,818 & 512 & $\geq7$ & $\geq3$ & 1 & C & Single & Aircraft dataset \\
MSAR~\cite{xia2022crtranssar,chen2022large} & 2022 & Det. & 28,499 & $256\sim2048$ & $\geq4$ & $\geq6$ & 1 & C & Quad & Terrestrial and maritime targets \\
OGSOD~\cite{wang2023category} & 2023 & Det. & 18,331 & 256 & $\geq3$ & $\geq2$ & 3 & C & Double & Targets include bridges, oil tanks, and harbours \\
\bottomrule
\end{tabular}}
\end{table*}

\subsection{Model Architecture}
Two model backbone types were considered for SAR target recognition. The first utilized a vision transformer (ViT)~\cite{dosoViTskiy2020image}, commonly used in SSL, offering good scalability of model parameters. The second architecture employed ConvNeXt-V2~\cite{woo2023convnext}, which offered the same scalability as ViT but maintained the efficiency of a convolutional neural network. In addition to the scalability of model parameters, remote sensing tasks also need to consider image properties. For example, SAR targets typically exhibit a small foreground and a dynamic context range. Swin transformers could outperform ViTs with a hierarchical structure yet are unsuitable when using drop patches in MIM to preserve computing resources. Therefore, we finally considered a variant of ViT, a hierarchical vision transformer (HiViT)~\cite{zhang2023hivit}, which improved the input spatial resolution and retained ViT properties for MIM.

\begin{figure*}[!tb]
\centering
\includegraphics[]{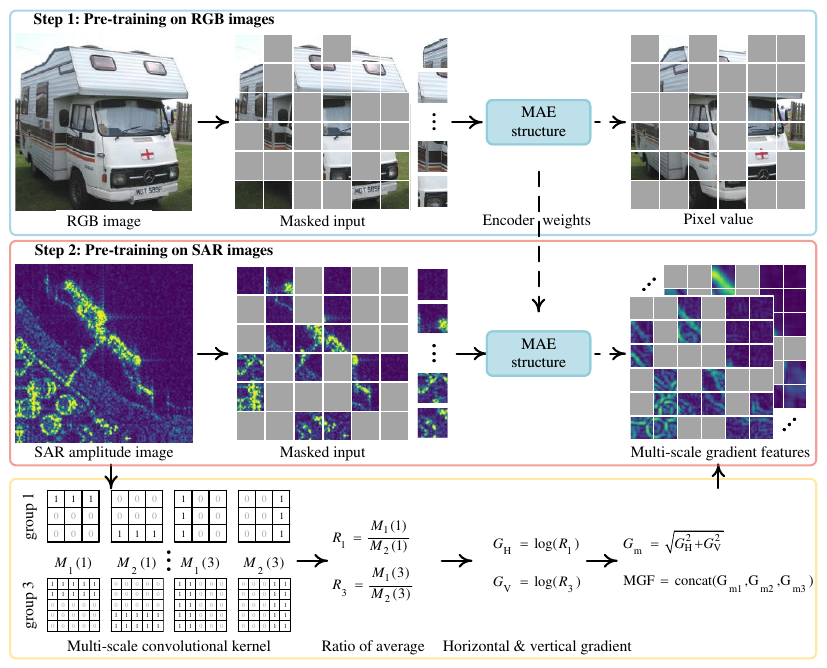}
\caption{Two-step pre-training process. The first involved performing MIM on ImageNet data to obtain better initialization weights for model diversity, as shown in Fig.~\ref{visual_attention_distance} (c). The second involved performing MIM on SAR images with high-quality guide signals \textcolor{black}{that are multi-scale gradient features suppressing speckle noise and extracting target edges}.}
\label{fig3_framework}
\end{figure*}

\subsection{Proposed Pre-training Method}
MIM was used as a pretext task for SSL pre-training, and masked autoencoders (MAE)~\cite{he2022masked} were employed to drop patches and preserve computational resources. MIM can help foundation models achieve SAR image interpretation by recognizing contextual relationships in objects (a key point when applying MIM). However, SAR utilizes a type of coherent imaging, which involves speckle noise that can interfere with a pretext task. As such, SARATR-X employs two pre-training steps to construct a foundation model, as shown in Fig.~\ref{fig3_framework}. The first step provides ImageNet weight to increase attention diversity and avoid the interference of SAR speckle noise in the early stages of the second step. We use multi-scale gradient features to suppress speckle noise throughout the second SAR pre-training step.

The first step involves performing MIM with ImageNet data to obtain better initialization weights because visible light images contain more and better signatures than SAR images. We simplified the multistage pre-training of MSFA, which performs SSL on ImageNet with a backbone, detection task pre-training on DOTA with a whole framework, and detection task finetuning on SAR images. SARATR-X uses the pre-training weights from ImageNet as initialization weights for the SAR pre-training step. This approach enhances the diversity of attention during SAR pre-training, as shown in Fig.~\ref{visual_attention_distance}. In contrast, random initialization leads to the convergence of attention toward the same pattern as in SAR pre-training with MAE. \textcolor{black}{Besides, the natural image weights also compensate well for the lack of diversity in the bottom layer of HiViT in SAR image pre-training and decreased diversity in the top layer due to MGFs.} The ImageNet pre-training backbone weights were obtained from an open source to reduce pre-training time. This process of using pre-trained ImageNet weights is called SSL-ImageNet \& SAR.
 
The second step involves performing MIM with SAR images. As mentioned previously, SAR image noise is a challenging problem that has been investigated in FG-MAR, SAR-JEPA, and MSFA, which have discussed features such as CannyEdge~\cite{canny1986computational}, HOG~\cite{dalal2005histograms}, Haar-like~\cite{viola2001rapid}, SAR-HOG~\cite{song2016sar}, and SAR-SIFT~\cite{dellinger2014sar}. Different feature combinations can be used to achieve the best results~\cite{li2024sardet100k}, but the simplest gradient features follow our previous SAR-JEPA approach to avoid excessive runtimes during complex feature selection. As such, MGFs~\cite{li2023self} were used to suppress speckle noise and extract target shapes.

\textbf{Multi-scale gradient feature -} 
\textcolor{black}{Multiplicative speckle noise causes distortions in the strong scattering of the region, which interferes with the SSL for the target features. Therefore, we would like to suppress the noise employing the target edge information for learning. However, Differential gradient have false points in strong target regions due to multiplicative speckle noise in SAR images~\cite{touzi1988statistical,bovik1988detecting}. This result is because multiplicative noise leads to scattering points with significant strong and weak amplitude variations in the strong scattering region, especially for strongly scattering metallic targets. Therefore, using a gradient ratio within a region can improve stability. It can reduce the interference in the strong and weak points inner target regions due to speckle noise. } In this study, MGF employed a gradient by ratio~\cite{song2016sar, dellinger2014sar} to obtain the relevant gradient features $G_\text{m}$: 
\begin{align}
\label{eq1}
R_i &= \frac{M_1(i)}{M_2(i)},\\
\label{eq2}
G_\text{H} &= log(R_1),\\
\label{eq3}
G_\text{V} &= log(R_3),
\end{align}
where $R_i$ denotes the average ratio in different directions. $M_1(i)$ and $M_2(i)$ denote area averages on opposite sides of a current pixel along the direction $i$, where $i = 1$ represents the horizontal direction and $i = 3$ indicates the vertical direction. The area averages can be calculated from an input image and four fixed convolution kernels (see Fig.~\ref{fig3_framework}). Eqs.~\ref{eq2} and~\ref{eq3} then use logarithms to perform the vertical gradient calculation~\cite{dellinger2014sar}, where $G_\text{H}$ is the horizontal gradient and $G_\text{V}$ is the vertical gradient. 
\begin{align}
\label{eq4}
G_\text{m} &= \sqrt{G_\text{H}^2+G_\text{V}^2}, \\
\label{eq5}
\text{MGF} &= \text{concat}(G_\text{m1}, G_\text{m2}, G_\text{m3}), 
\end{align}

Due to the dynamic range required for various targets in remote sensing~\cite{li2023large}, MGF is constructed with convolutional kernels of different sizes. We set the kernel scale $r$ equal to 9, 13, and 17 to obtain $G_\text{m1}$, $G_\text{m2}$, and $G_\text{m3}$, and the whole convolutional kernel size is odd square $2r+1$ to calculate the average ratio in four different directions. 

\subsection{Evaluation with Recognition Tasks}
Fine-grained classification datasets comprised of vehicles, ships, and aircraft were merged to form a new SAR classification dataset called SAR-VSA (Vehicles, Ships, and Aircraft) in Table~\ref{table_dataset}. \textcolor{black}{We aim to increase the number of classes and assess the proposed improvements in feature quality}. SAR-VSA compares SSL model performance with few-shot settings in Sec.~\ref{Experiments}. We then report the SARATR-X results for existing classification and detection settings, datasets, and algorithms in Sec.~\ref{Leveraging} \textcolor{black}{to show the powerful potential of a foundation model}.

\section{SARATR-X Experiments}
\label{Experiments}
We first performed SSL on the pre-training dataset without label information and then fine-tuned the pre-trained model on a classification dataset using few-shot classification tasks and linear probing settings to analyze improvements made to SARATR-X. We also discuss the scalability of the proposed technique. Pre-training was first performed on eight NVIDIA RTX3090 GPUs. The SAR pre-training SARDet-180K dataset consisted of 14 SAR datasets (see Table~\ref{table_dataset}). Specifically, a few-shot SAR classification dataset, SAR-VSA, included 25 fine-grained targets from three SAR datasets (MSTAR~\cite{10283916}, FUSAR-Ship~\cite{sun2022scan}, and SAR-ACD~\cite{sun2022scan}). It is difficult to ensure training convergence by fine-tuning whole model parameters in small sample cases. As such, we used linear probing~\cite{he2022masked}, which included a batch normalization layer, to adjust for differences in the statistical data properties and reduce the number of fine-tuning parameters. Detailed settings can be found in Appendix A. 

\begin{table*}[!tb]
\centering
\caption{Results for a classification dataset containing 25 categories. Linear probing was performed during few-shot SAR target classification. According to the results, we recommend using pre-trained ImageNet weights during initialization before pre-training the SAR datasets. We also recommend the HiViT model backbone for recognizing small targets in SAR images. Average accuracy was employed as the classification metric. \textbf{Bolded} text indicates the best result, while \underline{underlined} text is the next best result. SL: supervised learning. SSL: self-supervised learning.}
\label{table_result}
\renewcommand\arraystretch{1.2}
\begin{tabular}{cclcc p{0.6cm}<{\centering}p{0.6cm}<{\centering}p{0.6cm}<{\centering}} 
\toprule
\multirow{2}{*}{\textbf{Backbones}} & \multirow{2}{*}{\textbf{Params}} & \multicolumn{3}{c}{\textbf{Pre-Training}} & \multicolumn{3}{c}{\textbf{Classification (N-Shot)}} \\ 
\cmidrule(lr){3-5}\cmidrule(lr){6-8}
 & & \multicolumn{1}{c}{Settings} & Dataset & Method & 5 & 10 & 20 \\ 
\cmidrule(lr){1-8}
ConvNeXt-V2 & 89M & SL-ImageNet & ImageNet-1K & Supervised & 52.5 & 61.7 & 70.5 \\
ConvNeXt-V2 & 89M & SSL-ImageNet & ImageNet-1K & FCMAE & 47.2 & 54.5 & 64.0 \\
ConvNeXt-V2 & 89M & SSL-SAR & SAR images & FCMAE & 52.7 & 60.9 & 67.7 \\
ConvNeXt-V2 & 89M & SSL-ImageNet $\&$ SAR & ImageNet $\&$ SAR & FCMAE & 54.7 & 61.5 & 69.5 \\ 
\cmidrule(lr){1-8}
ViT & 86M & SL-ImageNet & ImageNet-1K & Supervised & 58.6 & 65.7 & 74.2 \\
ViT & 86M & SSL-ImageNet & ImageNet-1K & MAE & 50.7 & 58.0 & 65.5 \\
ViT & 86M & SSL-SAR & SAR images & MAE & 54.1 & 61.5 & 68.2 \\
ViT & 86M & SSL-ImageNet $\&$ SAR & ImageNet $\&$ SAR & MAE & 65.8 & 76.4 & 83.6 \\ 
\cmidrule(lr){1-8}
HiViT & 66M & SL-ImageNet & ImageNet-1K & Supervised & 49.0 & 55.8 & 63.3 \\
HiViT & 66M & SSL-ImageNet & ImageNet-1K & MAE & 53.0 & 60.3 & 69.3 \\
HiViT & 66M & SSL-SAR & SAR images & MAE & 64.9 & 72.7 & 79.9 \\
\underline{HiViT} & 66M & \underline{SSL-ImageNet $\&$ SAR} & ImageNet $\&$ SAR & \underline{MAE} & \underline{71.5} & \underline{78.5} & \underline{84.0} \\
\textbf{HiViT} & 66M & \textbf{SSL-ImageNet $\&$ SAR} & ImageNet $\&$ SAR & \textbf{Ours} & \textbf{76.5} & \textbf{80.8} & \textbf{85.1} \\
\bottomrule
\end{tabular}
\end{table*}

\begin{table}[!tb]
\centering
\caption{Comparisons of various target features for MIM. Many target features were found to be unsuitable for the multiplicative speckle noise in SAR images. These results inspired us to pursue different gradient features for SAR SSL. The pre-training settings in SSL-SAR and the model backbones formed the base version, while average velocity served as the classification metric. \textbf{Bolded} text indicates the best result, while \underline{underlined} text is the next best result.}
\label{table_result_SSL}
\renewcommand\arraystretch{1.2}
\begin{tabular}{cc p{0.6cm}<{\centering}p{0.6cm}<{\centering}p{0.6cm}<{\centering}} 
\toprule
\multirow{2}{*}{\textbf{Model}} & \multicolumn{1}{c}{\multirow{2}{*}{\textbf{Target Feature}}} & \multicolumn{3}{c}{\textbf{Classification (N-shot)}} \\ 
\cmidrule(lr){4-5} 
 & & 5 & 10 & 20 \\ 
\cmidrule(lr){1-5}
ViT & Pixel Value~\cite{he2022masked} & 54.1 & 61.5 & 68.2 \\
ViT & Low Pass Filter~\cite{liu2023pixmim} & 53.8 & 60.5 & 66.7 \\
ViT & HOG Feature~\cite{wang2023feature} & 39.7 & 48.7 & 56.6 \\
ViT & Deep Feature~\cite{assran2023self} & 27.8 & 35.6 & 41.7 \\
\cmidrule(lr){1-5}
HiViT & Pixel Value~\cite{he2022masked} & 64.9 & 72.7 & 79.9 \\
HiViT & HOG Feature~\cite{wang2023feature} & 58.2 & 64.8 & 71.7 \\
HiViT & \underline{SAR-HOG}~\cite{song2016sar} & \underline{75.1} & \underline{80.2} & \underline{83.9} \\
HiViT & \textbf{MGF} & \textbf{76.0} & \textbf{81.1} & \textbf{84.5} \\
\bottomrule
\end{tabular}
\end{table}

\subsection{Comparison of Model Backbones}
Table~\ref{table_result} compares different model backbones used for SAR ATR, including ConvNeXt-V2~\cite{woo2023convnext}, ViT~\cite{dosoViTskiy2020image}, and HiViT~\cite{zhang2023hivit}. The results indicated that ViT outperformed ConvNeXt-V2. On the one hand, ViT is more flexible than ConvNeXt-V2 for learning contextual information in SAR images. On the other hand, multiple downsampling steps in ConvNeXt-V2 resulted in the loss of small targets, while ViT maintained the same spatial resolution in different layers. \textcolor{black}{HiViT outperformed ViT and employed small (4 × 4) input patches}, capturing small target features well. Fig.~\ref{visual_attention_distance} demonstrates that HiViT also offered a superior variable attention distance compared to ViT due to the small target information common in remote sensing. Therefore, we use HiViT as the backbone of our SARATR-X.

\subsection{Strategy of two-step pre-training}
Here, we discuss the two-step pre-training strategy included to make full use of available model weights and SAR datasets. Table~\ref{table_result} includes four pre-training settings: SL-ImageNet, SSL-ImageNet, SSL-SAR, and SSL-ImageNet \& SAR. SL-ImageNet was pre-trained on ImageNet using supervised learning\footnote{We used open-source weights from GitHub to conduct the experiments. Supervised weights in ConvNeXt-V2 and HiViT were obtained with supervised fine-tuning after SSL.}; SSL-ImageNet was pre-trained on ImageNet from scratch using SSL; SSL-SAR was pre-trained from scratch on our SAR pre-training dataset; SSL-ImageNet \& SAR pre-trained the model on a SAR dataset based on initialized weights from SSL-ImageNet.

Notice the additional supervised information introduced by SL-ImageNet did not necessarily improve SAR ATR performance (\emph{e.g.}, the linear probing performance of SL-ImageNet for HiVit was lower than that of SSL-ImageNet). SSL-SAR achieved better results than SSL-ImageNet using less data (12\%), reflecting large differences in target features between the two images. However, ImageNet pre-training weights did provide a good initialization for lower features, such as shape and texture in SSL, with visible spectral remote sensing~\cite{10110958} and medical images~\cite{zhou2023foundation}. Our experiments also confirmed that using SSL-ImageNet as initialization weights improved the pre-training performance of SAR images (see Table~\ref{table_result}) and attention diversity (see Fig.~\ref{visual_attention_distance}). As such, SARATR-X employed the SSL-ImageNet \& SAR settings to complement the richness of pre-training.

\subsection{Design of target signals for SAR images}

After investigating the model backbone and learning strategy, we focused on target features for SSL methods used with SAR images. One key point for MIM is designing high-quality guide signals due to the unique multiplicative speckle noise in SAR images. This means we need to suppress noise and enhance target features. As seen in Table~\ref{table_result_SSL}, we considered five target features (pixel values)~\cite{he2022masked}, including a low pass filter~\cite{liu2023pixmim}, HOG features~\cite{wang2023feature}, deep features~\cite{assran2023self}, SAR-HOG~\cite{song2016sar}, and gradient by ratio~\cite{dellinger2014sar, song2016sar}). All SSL methods use the SSL-SAR setting and base version. We first considered whether existing methods based on ViT were suitable for SAR image classification. PixMIM~\cite{liu2023pixmim} applies a low pass filter to remove high-frequency components, driving the model to focus on shape information. However, PixMIM did not outperform MAE because the noise type in SAR is multiplicative, and the filter parameters require a trade-off between the target and noise. FG-MAE~\cite{wang2023feature} uses HOG to capture SSL features in SAR scene-level tasks, though we found that HOG did not ensure accurate SAR target features. Target regions typically exhibited strong scattering values, and the included speckle noise often caused the gradient computations to exhibit strong false points in these regions. In addition, I-JEPA~\cite{assran2023self} proposed deep networks for use as target feature encoders to capture deep semantic features. However, this can lead to training overfitting noise and a failure to effective features.

\begin{figure}[!tb]
\centering
\includegraphics[width=8.0cm]{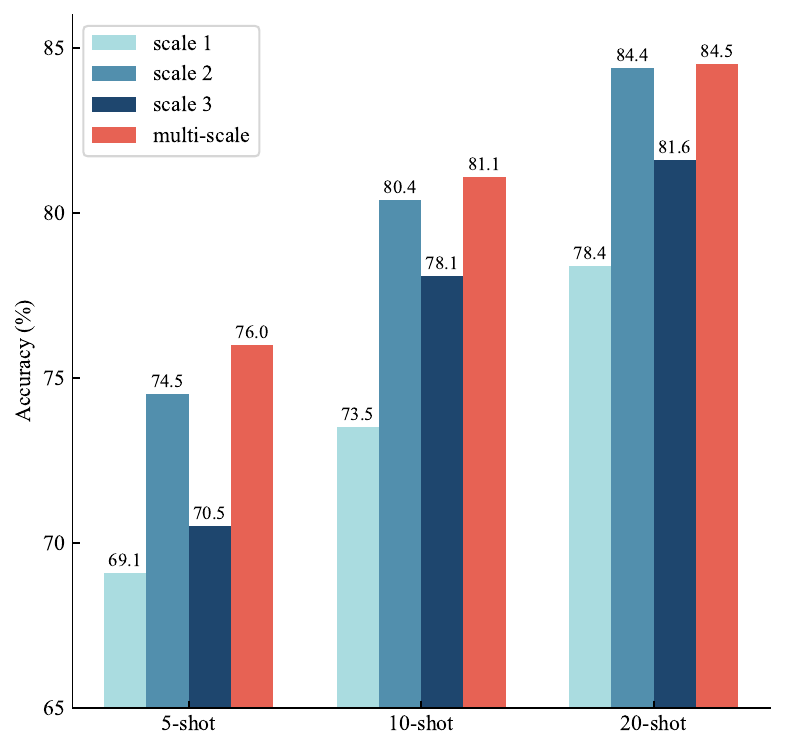}
\caption{Discussions of single and multi-scale kernel settings for MGF. Here, the scale 1/2/3 assumes $r$ equal to 9/13/17, as the multi-scale contacts all scales. This multi-scale approach is more suitable than a single-scale technique for various targets in remote sensing images.}
\label{fig_analysis_multiscale}
\end{figure}

As such, we chose SAR features as target features to enhance HiViT. SAR-HOG changes the gradient calculations for HOG features and uses gradient by ratio to solve for the speckle noise, thereby outperforming pixel values and HOG. Inspired by PixMIM, we prefer to directly use the target shape (\emph{i.e.}, gradient features) as a target feature~\footnote{SAR-HOG uses the same multi-scale settings to illustrate that simple gradient features can effectively represent target shape as an SSL guide signal.}. In addition, multi-scale methods can improve the feature representations of various small targets common in remote sensing. Discussions of kernel settings used for computing gradients are illustrated in Fig.~\ref{fig_analysis_multiscale}. Scale can also affect feature quality, as a smaller scale is finer for small target edge extraction, while a larger scale is more suitable for large targets and noise suppression. Therefore, combining features of different scales (see Fig.~\ref{fig_analysis_multiscale}) offers improvements for various image target sizes.

\subsection{Analysis}
As stated above, the primary contributions of SARATR-X can be summarized as follows: the HiViT architecture avoids the loss of small target information. SSL-ImageNet \& SAR use ImageNet pre-training weights to provide a good initialization for diversity perceptual capabilities. MGF ensures high-quality target features and suppresses speckle noise under SSL with SAR images. By taking advantage of these insights, SARATR-X can learn high-quality target features from noisy SAR remote sensing images, as seen in Table~\ref{table_result}. In the next section, we analyze the diversity and scalability of SARATR-X.

\begin{figure*}[!tb]
\centering
\includegraphics[width=17.3cm]{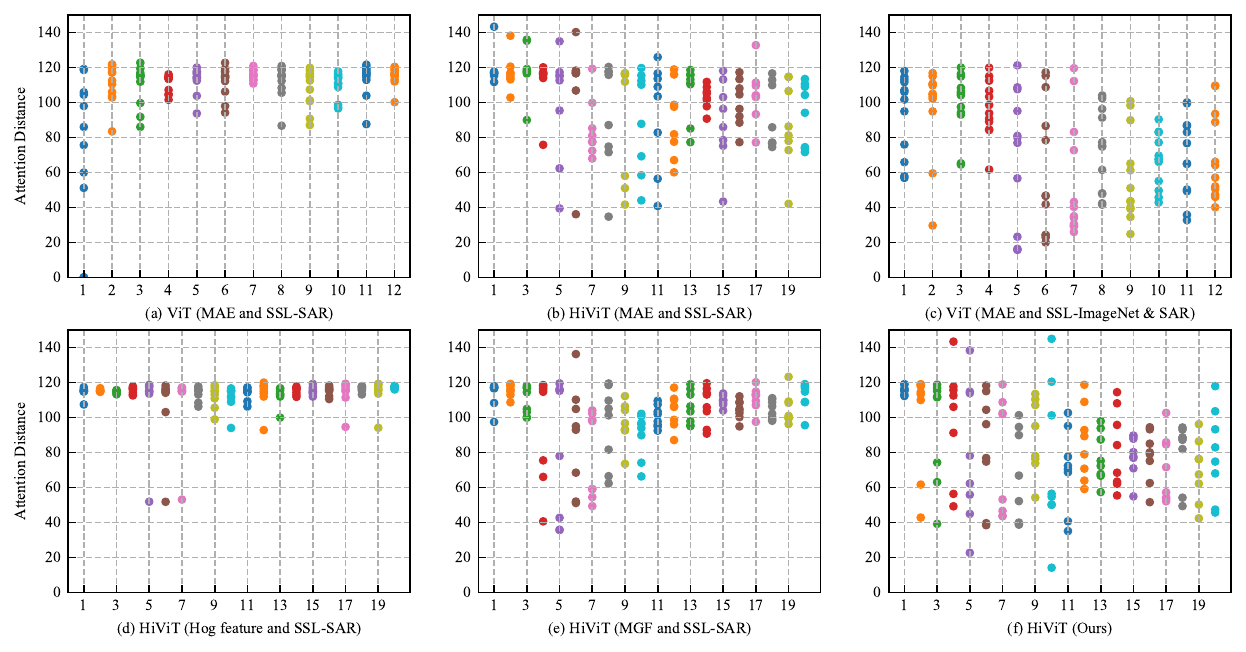}
\caption{Averaged attention distances for various attention heads (the x-axis is the attention head \emph{w.r.t} layer number, and point colors represent different layers for better visualization) in the SSL models. Attention distance represents the range of a receptive field. We focused specifically on model architectures (Fig. (a) \emph{v.s.} Fig. (b)), initialization weights (Fig. (a) \emph{v.s.} Fig. (c)), and SSL signals (Fig. (d) \emph{v.s.} Fig. (e)) to ensure diverse attention ranges for SAR target recognition, including the HiViT architecture, ImageNet weights, and SAR target features.}
\label{visual_attention_distance}
\end{figure*}

\begin{figure*}[!tb]
\centering
\includegraphics[width=17.5cm]{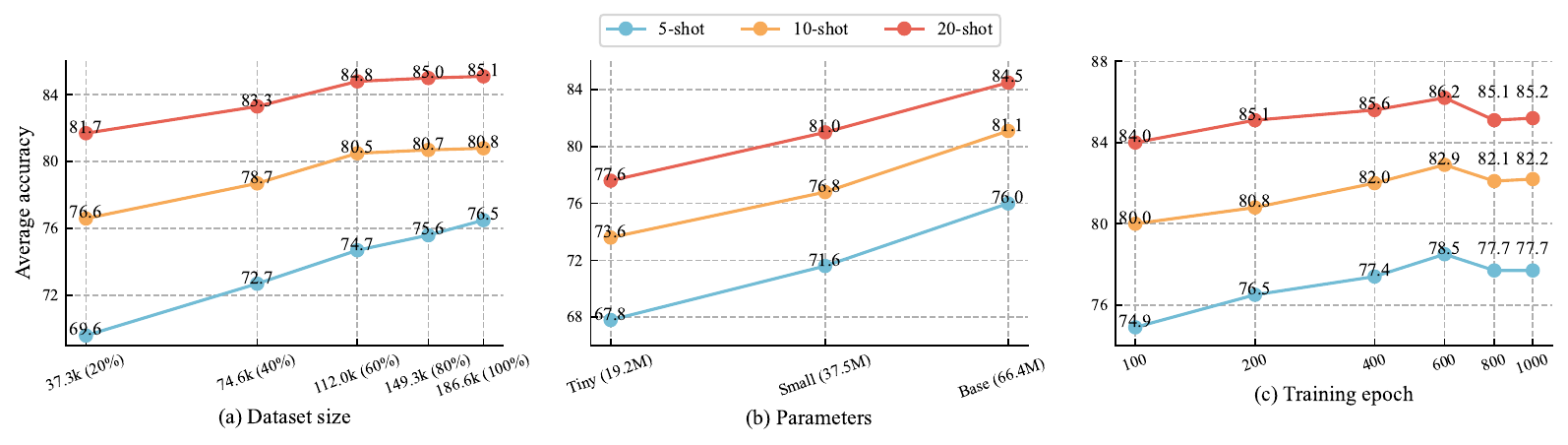}
\caption{\textcolor{black}{Scalability of SARATR-X} for dataset size, parameters, and training epochs with linear probing performance. While our method benefited from these three attributes, it is important to note that excessive training epochs often led to overfitting due to the dataset size.}
\label{fig_scalability}
\end{figure*}

\textbf{Visualization -}
Prior research~\cite{xie2023revealing} has shown that supervised pre-training and contrastive learning only model global information in higher layers, while MIM can model both local and global information. However, we observed that this effect was not only related to the chosen method but also to the data properties. Fig.~\ref{visual_attention_distance} (a) demonstrates that ViT with MAE focuses on global information due to large SAR image scenes, which differs from MIM modeling properties. As such, HiViT includes various attention distances from 40 to 140 in different layers with its high spatial resolution and hierarchical structure \textcolor{black}{but has decreased diversity in the bottom layer}, as seen in Fig.~\ref{visual_attention_distance} (b). In addition, using ImageNet weights for initialization solved this problem, as seen in Fig.~\ref{visual_attention_distance} (c). Similarly, Fig.~\ref{visual_attention_distance} (d) demonstrates that HOG features enhanced the noise interference, which limited feature diversity. MGF effectively extracts shape information from targets, focusing the model on diverse edge information in the lower layer. However, this approach removes textures and preserves edge information, which motivates higher layers to rely less on texture details and diminishes the attention range shown in Fig.~\ref{visual_attention_distance} (e). Thus, we combined SSL-ImageNet $\&$ SAR with MGF for two-step pre-training in Fig.~\ref{visual_attention_distance} (f). \textcolor{black}{However, the benefits of different modalities, especially a sufficient number of visible images, for SAR data need to be explored more. In the future, we plan to collect the satellite and airborne multimodal paired data and use various features, including filter pixels, handmade features, and learned features, to ensure the migration of the underlying feature extraction and the higher-level semantic information. Continuous learning~\cite{mendieta2023towards, huang2024generic} is also needed to prevent catastrophic forgetting.}

\begin{table}[!tb]
\centering
\caption{SAR classification results for the MSTAR SOCs and EOCs settings. SOCs exhibited a similar distribution and included ten fine-grained classes of vehicles. EOCs exhibited variations for imaging conditions that belonged to robustness classification and out-of-distribution problems. The EOCs used depression angle, configuration, and version variations to test robustness. We observed that self-supervised (BIDFC and Ours) and semi-supervised (EUAPS) performance significantly improved for few-shot tasks with additional unlabeled data, demonstrating the importance of the data. In addition, our method is robust to variations in imaging parameters. Accuracy$\uparrow$ was used as an evaluation metric, with detailed settings following the $N$-shot~\cite{zhang2021domain}. \textbf{Bolded} text indicated the best result, while \underline{underlined} text denoted the next best result. Detailed results are provided in Appendix B.3.}
\label{table_result_classification}
\renewcommand\arraystretch{1.2}
\resizebox{0.9\linewidth}{!}{%
\begin{tabular}{lc p{1.5cm}<{\centering}p{1.5cm}<{\centering}p{1.5cm}<{\centering}}
\toprule
\multicolumn{5}{c}{SOCs-Standard operating conditions (10-way)} \\
\cmidrule(){1-5}
\multicolumn{1}{c}{Method} & Year & 1-shot & 2-shot & 5-shot \\
\cmidrule(lr){1-5}
DKTS-N~\cite{zhang2021domain} & 2021 & 49.3 & 58.5 & 72.3 \\
ConvT~\cite{Wang2022global} & 2022 & 42.6 & 54.4 & 75.2 \\
HDLM~\cite{Wang2022recognition} & 2022 & - & - & 72.4 \\
BIDFC~\cite{zhai2022weakly} & 2022 & \underline{80.7} & \underline{85.3} & \underline{90.3} \\
CRID~\cite{wang2023crucial} & 2023 & 48.3 & 51.0 & 73.3 \\
EUAPS~\cite{10138441} & 2023 & - & - & 88.7 \\
PD~\cite{zhang2024optimal} & 2024 & 46.7 & 58.9 & 70.2 \\
\textbf{SARATR-X} & 2024 & \textbf{85.2} {\scriptsize (+4.5)} & \textbf{91.4} {\scriptsize (+6.1)} & \textbf{95.9} {\scriptsize (+5.6)}\\
\cmidrule(){1-5}
\multicolumn{5}{c}{EOCs-Depression angle variations (4-way)} \\
\cmidrule(){1-5}
\multicolumn{1}{c}{Method} & Year & 1-shot & 2-shot & 5-shot \\
\cmidrule(lr){1-5}
DKTS-N~\cite{zhang2021domain} & 2021 & 61.9 & 63.9 & 67.4 \\
ConvT~\cite{Wang2022global} & 2022 & 59.6 & \underline{64.1} & 68.2 \\
CRID~\cite{wang2023crucial} & 2023 & \underline{62.1} & 62.3 & \underline{74.5} \\
\textbf{SARATR-X} & 2024 & \textbf{93.4} {\scriptsize (+31.3)} & \textbf{97.3} {\scriptsize (+33.2)} & \textbf{98.9} {\scriptsize (+24.8)}\\
\cmidrule(){1-5}
\multicolumn{5}{c}{EOCs-Target configuration variations (4-way)} \\
\cmidrule(){1-5}
\multicolumn{1}{c}{Method} & Year & 1-shot & 2-shot & 5-shot \\
\cmidrule(lr){1-5}
DKTS-N~\cite{zhang2021domain} & 2021 & 47.3 & 53.6 & 62.2 \\
ConvT~\cite{Wang2022global} & 2022 & 44.3 & 51.9 & 64.1 \\
CRID~\cite{wang2023crucial} & 2023 & \underline{62.8} & \underline{65.7} & \underline{74.1} \\
\textbf{SARATR-X} & 2024 & \textbf{65.0} {\scriptsize (+2.2)} & \textbf{74.0} {\scriptsize (+8.3)} & \textbf{78.3} {\scriptsize (+4.2)}\\
\cmidrule(){1-5}
\multicolumn{5}{c}{EOCs-Target version variations (4-way)} \\
\cmidrule(){1-5}
\multicolumn{1}{c}{Method} & Year & 1-shot & 2-shot & 5-shot \\
\cmidrule(lr){1-5}
DKTS-N~\cite{zhang2021domain} & 2021 & 48.9 & 55.1 & 65.6 \\
ConvT~\cite{Wang2022global} & 2022 & 42.3 & \underline{58.3} & \underline{68.1} \\
CRID~\cite{wang2023crucial} & 2023 & \underline{53.5} & 56.2 & 67.2 \\
\textbf{SARATR-X} & 2024 & \textbf{65.3} {\scriptsize (+11.8)} & \textbf{76.5} {\scriptsize (+20.3)} & \textbf{82.8} {\scriptsize (+15.6)}\\
\bottomrule
\end{tabular}}
\end{table}

\begin{table}[!tb]
\centering
\caption{SAR detection results for SARDet-100K, OGSOD, SSDD, and SAR-Aircraft. Our proposed SARATR-X achieved competitive performance for various detection datasets. mAP$\uparrow$ was used as an evaluation metric. \textbf{Bolded} text indicates the best result, while \underline{underlined} text denotes the next best result. Detailed results are in Appendix B.3.}
\label{table_result_detection}
\renewcommand\arraystretch{1.2}
\begin{threeparttable}
\resizebox{0.9\linewidth}{!}{%
\begin{tabular}{l c p{1.2cm}<{\centering} p{1.2cm}<{\centering} p{1.2cm}<{\centering}} 
\toprule
\multicolumn{5}{c}{SARDet-100K (Object detection)} \\
\cmidrule(){1-5}
\multicolumn{1}{c}{\multirow{1}{*}{Method}} & Year & $\rm{mAP}\uparrow$ & $\rm{mAP_{50}}\uparrow$ & $\rm{mAP_{75}}\uparrow$ \\
\cmidrule(lr){1-5 } 
Deformable {DETR}~\cite{zhu2020deformable} & 2020 & 50.0 & 85.1 & 51.7 \\
Swin Transformer~\cite{liu2021swin} & 2021 & 53.8 & 87.8 & 59.0 \\
VAN~\cite{guo2023visual} & 2022 & 53.5 & 86.8 & 58.0 \\
ConvNeXt~\cite{liu2022convnet} & 2022 & 55.1 & 87.8 & 59.5 \\
MSFA~\cite{li2024sardet100k} & 2024 & \underline{56.4} & \underline{88.2} & \underline{61.5} \\
\textbf{SARATR-X} & 2024 & \textbf{57.3} {\scriptsize (+0.9)} & \textbf{88.7} {\scriptsize (+0.5)} & \textbf{62.8} {\scriptsize (+1.3)} \\
\cmidrule(){1-5}
\multicolumn{5}{c}{OGSOD (Object detection)} \\
\cmidrule(){1-5}
\multicolumn{1}{c}{\multirow{1}{*}{Method}} & Year & $\rm{mAP}\uparrow$ & $\rm{mAP_{50}}\uparrow$ & $\rm{mAP_{75}}\uparrow$ \\
\cmidrule(lr){1-5} 
Generalized Focal~\cite{he2019bounding} & 2019 & 41.8 & 67.6 & - \\
Sparse R-CNN~\cite{sun2021sparse} & 2021 & 38.7 & 65.6 & - \\
Object Box~\cite{zand2022objectbox} & 2022 & 40.1 & 76.6 & - \\
YOLOv7~\cite{wang2023yolov7} & 2022 & \underline{45.1} & \underline{79.2} & - \\
\textbf{SARATR-X} & 2024 & \textbf{52.0} {\scriptsize (+6.9)} & \textbf{85.9} {\scriptsize (+6.7)} & \textbf{51.3}\\
\cmidrule(){1-5}
\multicolumn{5}{c}{SSDD (Ship detection)} \\
\cmidrule(){1-5}
\multicolumn{1}{c}{\multirow{1}{*}{Method}} & Year & $\rm{AP}\uparrow$ & $\rm{AP_{50}}\uparrow$ & $\rm{AP_{75}}\uparrow$ \\
\cmidrule(lr){1-5 } 
FBR-Net~\cite{fu2021anchor} & 2021 & - & 94.1 & 59.1 \\
CenterNet++~\cite{guo2021centernet++} & 2021 & - & 95.1 & - \\
CRTransSar~\cite{xia2022crtranssar} & 2022 & - & 97.0 & \underline{76.2} \\
YOLO-Lite~\cite{ren2023yolo} & 2023 & - & 94.4 & - \\
FEPS-Net~\cite{bai2023feature} & 2023 & 59.9 & 96.0 & 67.5 \\
$\rm{CS}^n$Net~\cite{Chen2023CSnNet} & 2023 & \underline{64.9} & \underline{97.1} & - \\
\textbf{SARATR-X} & 2024 & \textbf{67.5} {\scriptsize (+2.6)} & \textbf{97.3} {\scriptsize (+0.2)} & \textbf{83.5} {\scriptsize (+7.3)} \\
\cmidrule(){1-5}
\multicolumn{5}{c}{SAR-Aircraft (Aircraft detection)} \\
\cmidrule(){1-5}
\multicolumn{1}{c}{\multirow{1}{*}{Method}} & Year & $\rm{mAP}\uparrow$ & $\rm{mAP_{50}}\uparrow$ & $\rm{mAP_{75}}\uparrow$ \\
\cmidrule(lr){1-5} 
Cascade R-CNN~\cite{cai2018cascade} & 2018 & - & 75.7 & 58.9 \\
RepPoints~\cite{yang2019reppoints} & 2019 & - & 72.6 & 53.3 \\
SKG-Net~\cite{fu2021scattering} & 2021 & - & 70.7 & 46.4 \\
SA-Net~\cite{wang2023sar} & 2023 & - & \underline{77.7} & \underline{62.8} \\
\textbf{SARATR-X} & 2024 & \textbf{58.7} & \textbf{86.1} {\scriptsize (+5.7)} & \textbf{64.7} {\scriptsize (+3.3)} \\
\bottomrule
\end{tabular}}
\end{threeparttable}

\end{table}

\textbf{Scaling experiment -}
Although MIM learns effectively and scales with data and model resources~\cite{xie2023data}, a question arises as to whether our method can ensure scalability for MIM when dealing with noisy data, such as SAR. Fig.~\ref{fig_scalability} presents the results of a scaling experiment from three perspectives: dataset size, parameters, and training epochs. 
Despite our pre-training set comprising 186,660 images, which is smaller than ImageNet-1K, we observed a significant rising curve in downstream task performance with increasing data and parameter quantities in Figs.~\ref{fig_scalability} (a) and (b). This result indicated that the foundational model could fully achieve its potential in SAR images by extracting high-quality features as guiding signals. However, as in~\cite{xie2023data}, the model tended to overfit during extended training epochs when the pre-training set contained about 100,000 images in ImageNet. In addition, SAR image noise and low resolution further aggravated the overfitting. Regardless, SARATR-X outperformed our previous study (SAR-JEPA), which \textcolor{black}{overfitted} at 400 epochs with 94,776 SAR images. Thus, there is a need to continue investigating new ways to ensure high-quality feature representations when extending SAR foundation models.

\section{Leveraging SARATR-X for Recognition}
\label{Leveraging}
\begin{figure*}[!tb]
\centering
\includegraphics[width=16.5cm]{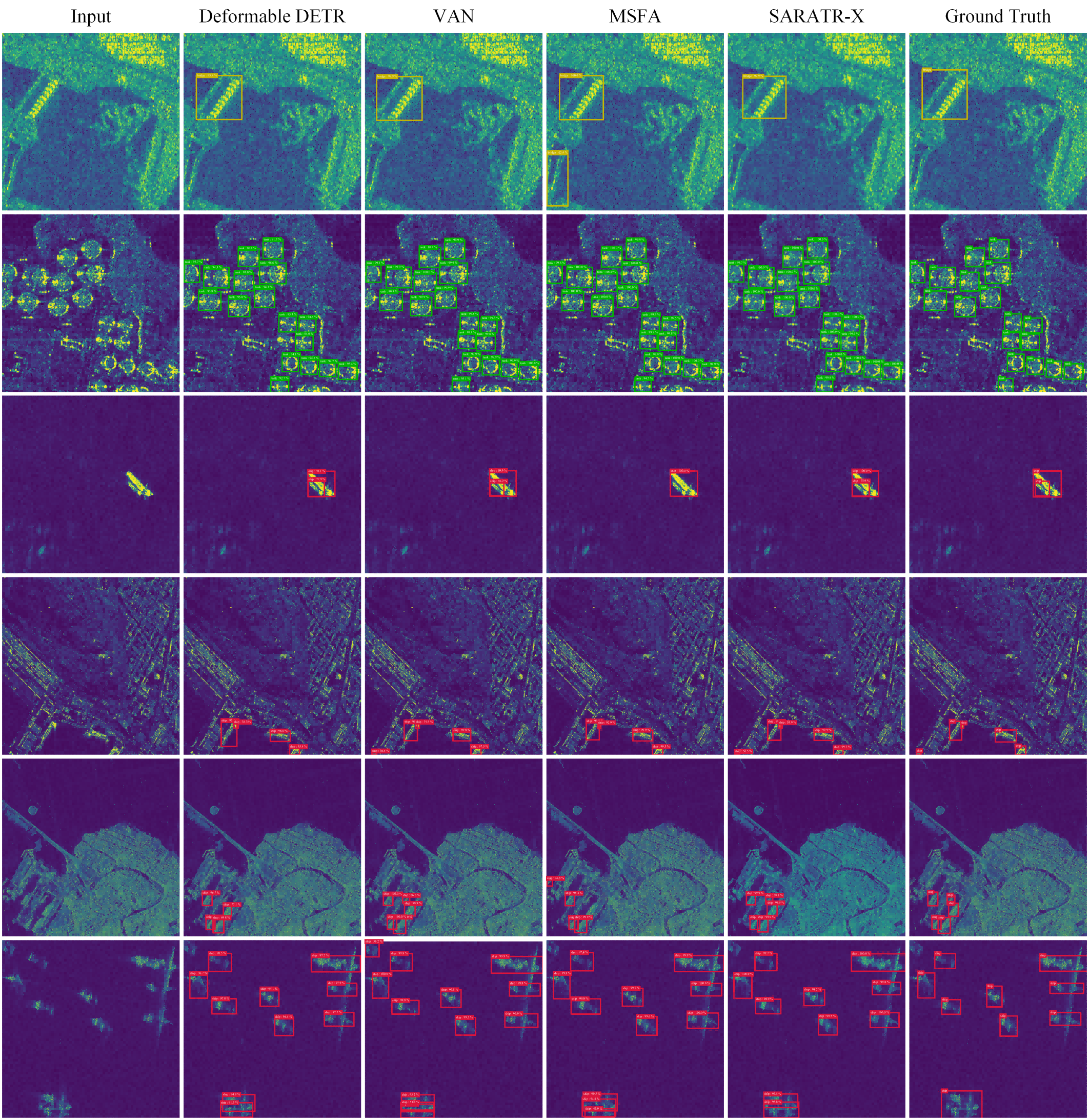}
\caption{\textcolor{black}{Visualization of detection on SARDet-100K. False alarms and missed detections are common in SAR images, especially when similar targets overlap and complex scenes are present. While our method effectively improves the detection effect by learning the contextual information in the image, target detection in complex scenes and low-quality images is still very difficult.} }
\label{fig_visual}
\end{figure*}

We have discussed different aspects of SARATR-X, but there are many datasets and specialized models available for SAR ATR. Therefore, we compared our proposed SARATR-X algorithm with other state-of-the-art techniques, such as supervised learning ($\rm{CS}^n$Net~\cite{Chen2023CSnNet} and PD~\cite{zhang2024optimal}), semi-supervised learning (EUAPS~\cite{10138441}), and self-supervised learning (MSFA~\cite{li2024sardet100k} and BIDFC~\cite{zhai2022weakly}). We focused on SAR recognition tasks, including image classification and object detection in Fig.~\ref{fig_sota}. More detailed settings\footnote{We removed the test sets of downstream tasks from the pre-training set samples, \textcolor{black}{and other method results are from their original article tables.}} are provided in Appendix B.

\textbf{Classification task -} 
Table~\ref{table_result_classification} details the performance of SARATR-X for the MSTAR~\cite{MSTAR} dataset with standard operating conditions (SOCs) and extended operating conditions (EOCs). Notice that SSL (BIDFC and ours) and semi-supervised models (EUAPS) significantly outperformed other methods for small samples with additional unlabeled data. Our results also surpassed the previous best by \emph{large margins}. For example, SOC 1-shot accuracy increased by 4.5\% and EOCs 1-shot accuracy increased by 15.1\% on average. Demonstrating the value of FMs in an era of rapidly growing SAR data. In particular, SARATR-X exhibited robustness to the EOC setting of variable imaging conditions. This result indicated that the foundation model could learn stable features and relationships from diverse imaging conditions in a large number of samples.

\textbf{Detection task -} 
As illustrated in Table~\ref{table_result_detection}, we reported a box mAP for SAR target detection with a horizontal bounding box for multi-category (SARDet-100K and OGSOD), ship (SSDD), and aircraft detection (SAR-Aircraft). SARATR-X outperformed our previous MSFA by 0.8 points on SARDet-100K. MSFA also includes complex training processes and target features, which employ multi-stage training between RGB and SAR images with three different target features (HOG~\cite{dalal2005histograms}, Haar-like~\cite{viola2001rapid}, WSTG~\cite{mallat2012group}), including a detection pre-training step. SARATR-X is thus simpler yet more effective for SAR images. \textcolor{black}{The detection visualization is shown in Fig.~\ref{fig_visual}, and we have fewer missed detections and false alarm results.} Notably, SARATR-X outperformed or offered comparable performance for multiple datasets, compared to several specifically designed detection methods shown in Table~\ref{table_result_detection}. 

Of course, our study is only a preliminary exploration of SSL for SAR image interpretation. More effective target features could be achieved in a data-knowledge dual-driven manner by further mining information on SAR imaging mechanisms and properties. Furthermore, given a larger dataset and computing power, the path of FMs will hopefully lead to generalized SAR interpretation, including target recognition, scene classification, semantic segmentation, and change detection, but this will require additional research.

\section{Conclusion and Future Perspectives}
\label{Conclusion}
This study proposed \textcolor{black}{a foundation model SARATR-X for SAR ATR}. First, a pre-training dataset SARDet-180K was constructed from 14 open-source datasets, including various targets, scenes, and sensors. The foundation model's pre-training backbone, SSL methods, and downstream tasks were then discussed in detail. Importantly, SARATR-X demonstrated superior performance on different target recognition datasets, demonstrating the potential of FMs in this field. We believe that further research on SAR foundation models, including SARATR-X, has the potential to generalize feature representations of SAR images and benefit all-day, all-weather target recognition in Earth observations. However, FMs research requires large data and SAR images are expensive and require specific imaging equipment and algorithms. Privacy and security also prevent the data from becoming open source. Therefore, we are particularly grateful to the publishers of open-source SAR target datasets. By making SARATR-X publicly available, we aim to accelerate the FMs in SAR target recognition by enabling researchers to use our code and weight to design better methods and explore downstream applications. 

\begin{figure}[!tb]
\centering
\includegraphics[width=7cm]{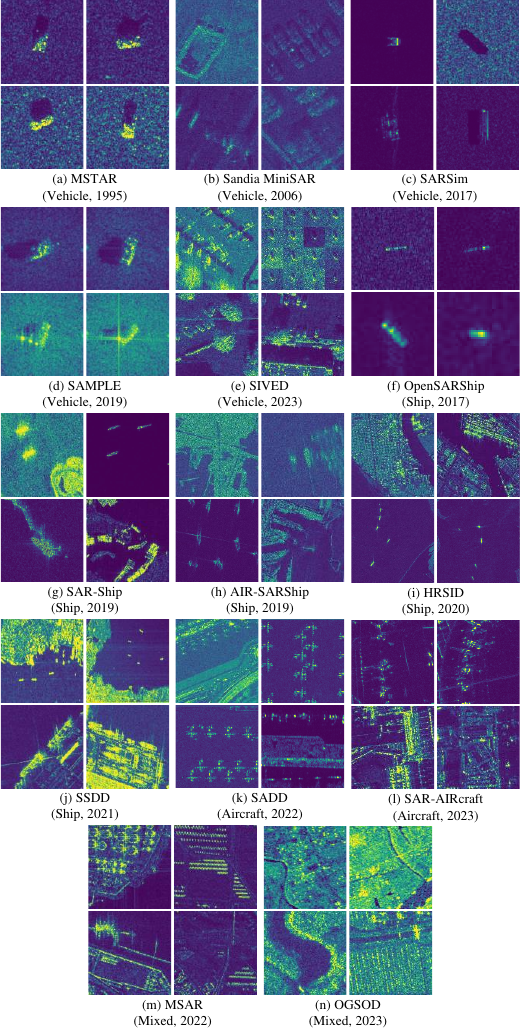}
\caption{Visualization of the fourteen datasets included in our pre-training. In this way, with the contribution of open-source researchers and the community, a pre-trained dataset for SAR ATR can be built with multiple targets, scenes, and sensors.}
\label{fig4_datasets}
\end{figure}

Although this work performs systematical investigations, several limitations and challenges will require exploration in a future study. The SAR images were derived from open-source SAR datasets, and the targets primarily included vehicles, ships, aircraft, oil tanks, etc. Thus, collecting target samples from increasingly unlabeled imagery could further expand the amount of data to reach a million level and the range of downstream applications. \textcolor{black}{The pre-training set image size~\cite{guo2025llava} and automatic target slicing will become important issues that need to be investigated due to the larger image size and contextual range of remote sensing images compared to natural images.} In terms of model architecture, learning the dynamic contextual information on space-time is also an important issue. In addition to target shape features, various scenarios and target signatures for SAR self-supervised learning need to be explored. Finally, investigating expert knowledge with text used for multimodal interactions and describing the relationship between targets and scenes could enhance the representation capabilities of FMs for SAR ATR. 

In conclusion, we have demonstrated the ability of SARATR-X to adapt to diverse SAR target datasets, achieving high performance and generalizability in classification and detection tasks. By taking full advantage of the rapid growth of SAR images, this SSL-based foundation model opens the door to generalized SAR target recognition.



\appendices

\section{Implementation Details for Section 4 and SARATR-X Experiments}
\label{Implementation Details}
Here are the details of the dataset and training settings.

\subsection{Pre-traing dataset setting}
\label{Pre-traing dataset setting}
we chose the most open-source dataset as pre-training. As shown in Fig.~\ref{fig4_datasets}, we collect data from open-source datasets based on our previous research~\cite {li2023self, li2024sardet100k}. Now, our pre-training dataset contains 14 open-source SAR target datasets. Here are brief descriptions of each dataset's targets, scenes, and sensors.

\begin{figure*}[!tb]
\centering
\includegraphics[width=\textwidth]{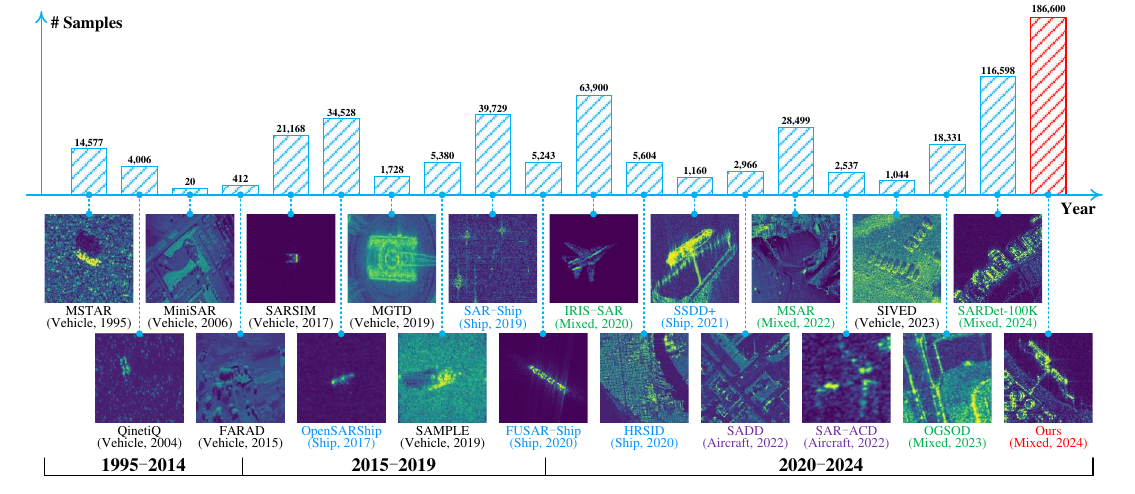}
\caption{Timeline of measured and simulated datasets used for SAR target recognition (SAR magnitude images were processed in pseudo-color for better visualization). SAR target datasets have increased rapidly since 2020, greatly enriching the diversity of available targets, scenes, and sensors. The different colours represent the target categories included in the dataset. Target categories primarily include fine-grained vehicles, ships, and aircraft. A mixed sample implies the target exhibits more than one category, such as the combination of various vehicles, ships, and aircraft in IRIS-SAR. These scenes may also contain cities, harbors, airports, etc. The sensor devices offer different resolutions and bands for satellite or airborne platforms. However, the number of samples (No. samples) in most datasets is only a few thousand across several categories, due to high costs and the difficulty of annotation. This inspired us to propose SARATR-X, with an integrated pre-training dataset SARDet-180K.}
\label{fig1_dataset}
\end{figure*}

\textbf{MSTAR}~\cite{MSTAR} is the most commonly used target classification dataset released by the Defense Advanced Research Projects Agency, USA. Its sensor is an X-band radar with HH polarization mode and 0.3 m resolution. It contains ten categories of military vehicles with various imaging angles, target variants, and other conditions but with simple grass scenes and not completely open source. 

\textbf{Sandia MiniSAR}~\cite{Sandia} is a 0.1 m resolution dataset based on a Ku-band airborne platform released by Sandia National Laboratories. The dataset contains scenes and targets such as aircraft on tarmacs, buildings in urban areas, and vehicles in desert areas but lacks official annotations.

\textbf{SARSim}~\cite{malmgren2017improving,kusk2016synthetic} is a fine-grained vehicle dataset created by Terma A/S, Denmark. The simulation system used for this dataset can generate X-band SAR images with resolutions ranging from 0.1m to 0.3m from CAD models. SARSim provides 21,168 vehicle samples in 7 categories (truck, car, motorbike, bus, tank, bulldozer, and pickup) and 3 scenes (grass, roads, and a mean of the two) with 7 imaging depression angles.

\textbf{SIVED}~\cite{lin2023sived} is a vehicle detection dataset with rotatable bounding box. It consists of vehicle slices from the MSTAR dataset~\cite{MSTAR} and vehicles in urban areas from the Sandia MiniSAR and FARAD datasets~\cite{Sandia}, and scenes include car parks, buildings, trees, roads, and others.

\textbf{SAMPLE}~\cite{lewis2019sar} is a synthetic and measured paired fine-grained vehicle dataset released by the Air Force Research Laboratory, USA. This dataset is simulated in X-band and 0.3 m resolution. The public version provides 5,380 images of ten categories of vehicle targets at partial imaging angles. 

\textbf{OpenSARShip}~\cite{li2017opensarship} is a ship slices dataset based on the European C-band Sentinel-1 satellite. Its resolution is 2.3 m to 17.4 m with VV and VH polarization. The dataset contains many ship slices from 10 busy ports. It has a diverse range of ship types but a significant category imbalance.

\textbf{SAR-Ship}~\cite{ref54} is a ship target detection dataset in complex scenes based on Chinese Gaofen-3 and European Sentinel-1 satellites. The public version of this dataset contains 39,729 images from two satellites in different imaging modes and resolutions. The dataset provides ship targets of various sizes in complex ocean scenes such as nearshore, distant seas, harbors, and islands.

\textbf{AIR-SARShip}~\cite{xian2019air} is a ship detection dataset based on the Chinese C-band Gaofen-3 satellite. AIR-SARShip-1.0 and AIR-SARShip-2.0 include 318 VV-polarised images with 1 and 3 m resolutions. This dataset includes harbors, islands, and different conditions of sea surfaces and covers thousands of ships.

\textbf{HRSID}~\cite{wei2020hrsid} is a high-resolution dataset for ship detection and instance segmentation based on the European C-band Sentinel-1B, German X-band TerraSAR-X and TanDEM-X satellites. HRSID consists of 5,604 cropped SAR images with 0.5 to 3 m resolutions. The scene is a busy area of maritime transport, such as harbors and estuarine cities, and the annotation targets are civil ships of different sizes.  

\textbf{SSDD}~\cite{zhang2021sar} is a commonly used SAR ship detection dataset. It is constructed based on Canadian RadarSat-2, German TerraSAR-X, and European Sentinel-1 satellites and contains different scenarios for the inshore and offshore of China and India. The dataset covers various ship sizes in different oceanic conditions with diverse clutter and noise interference. 

\textbf{SADD}~\cite{zhang2022sefepnet} is an aircraft detection dataset collected from the German X-band TerraSAR-X satellite. Its resolution is 0.5 m to 3 m with HH polarization. The dataset contains densely parked aircraft of different sizes on airport tarmacs, runways, and airport cemeteries. It has a large number of small-sized planes as well as the airport perimeter area.

\textbf{SAR-AIRcraft}~\cite{wang2023sar} is a aircraft detection dataset based on the Chinese C-band Gaofen-3 satellite with 1 m resolution and single polarization. The dataset collects seven types of aircraft of different sizes from three civil airports. It can support fine-grained aircraft detection and classification studies.

\textbf{MSAR}~\cite{xia2022crtranssar,chen2022large} is a multi-class target detection dataset based on the Chinese C-band HISEA-1 satellite in large-scale scenes. MSAR comprises 28,449 image slices with quad polarization and 1 m resolution. Scenes covered include airports, harbors, nearshore, islands, distant seas, and urban areas. The labeled target categories include aircraft, oil tanks, bridges, and ships. 

\textbf{OGSOD}~\cite{wang2023category} is a city object detection dataset collected from the Chinese C-band Gaofen-3 satellite with VV and VH polarization modes, and its resolution is 3 m. This dataset also contains optical images from Google Earth with 10 m resolution. It is annotated with static objects, including bridges, harbors, and oil tanks in urban areas.

\subsection{Classification dataset for performance Test}
\label{Classification task datasets}

\begin{table}
\centering
\caption{Description of our SAR classification dataset, named SAR-VSA, which contains 25 target classes. \# Train: Number of train samples. \# Test: Number of test samples.}
\label{table_class_dataset}
\renewcommand\arraystretch{1.2}
\begin{tabular}{lcc} 
\toprule
Fine-grained~category & \#~Train & \# Test \\ 
\cmidrule(lr){1-3}
anti-aircraft (ZSU234) & 299 & 274 \\ 
bulldozer (D7) & 299 & 274 \\ 
howitzer (2S1) & 299 & 274 \\
infantry vehicle (BMP2) & 698 & 587 \\
main battle tank (T62) & 299 & 273 \\
main battle tank (T72)  & 691 & 582 \\
patrol car (BRDM2) & 298 & 274 \\
personnel carrier (BTR60) & 256 & 195 \\
personnel carrier (BTR70) & 233 & 196 \\
truck (ZIL131) & 299 & 274 \\
\cmidrule(lr){1-3}
bridge & 1,023 & 438 \\
coastal land & 707 & 303 \\
land patch & 1,137 & 487 \\
sea clutter wave & 1,378 & 590 \\
sea patch & 1,250 & 535 \\
ship (cargo) & 366 & 156 \\
ship (fishing) & 248 & 106 \\
ship (tanker) & 150 & 64 \\ 
ship (others) & 312 & 133 \\
strong false alarms & 299 & 128 \\
\cmidrule(lr){1-3}
aircraft (Airbus A220) & 91 & 373 \\
aircraft (Airbus A330) & 97 & 415 \\
aircraft (Comac ARJ21) & 103 & 411 \\
aircraft (Boeing 737) & 100 & 428 \\
aircraft (Boeing 787) & 113 & 391 \\
\bottomrule
\end{tabular}
\end{table}

\begin{table}[!tb]\centering
\caption{Pre-training setting.}
\label{table: hyperparameter settings pre-training}
\renewcommand\arraystretch{1.2}
\begin{tabularx}{0.40\textwidth}{ p{3cm} L }
\toprule
Config & Value \\ 
\cmidrule(lr){1-2}
optimizer & AdamW~\cite{loshchilov2019decoupled} \\
base learning rate & 1.5e-4 \\
weight decay & 0.05 \\
optimizer momentum & \textit{}$\beta_1,\beta_2 = 0.9,0.95$~\cite{chen2020generative} \\
batch size & 800 \\
epoch & 200 \\
learning rate schedule & cosine decay~\cite{loshchilov2017sgdr} \\
warmup epoch~\cite{goyal2017accurate} & 5 \\
augmentation & ResizedCrop, HFlip, ColorJitter \\
\bottomrule
\end{tabularx}
\end{table}

\begin{table}[!tb]\centering
\caption{Classification settings (fine-tuning and linear probing).}
\label{table: hyperparameter settings classification}
\renewcommand\arraystretch{1.2}
\begin{tabularx}{0.40\textwidth}{ p{3cm} L }
\toprule
Config & Value \\ 
\cmidrule(lr){1-2}
optimizer & AdamW~ \\
base learning rate & 1e-3 \\
weight decay & 1e-4 \\
optimizer momentum & \textit{}$\beta_1,\beta_2 = 0.9,0.95$~ \\
batch size & 25 \\
epoch & 30 \\
learning rate schedule & cosine decay~ \\
warmup epoch & 1 \\
warmup type & constant \\
warmup learning rate & 1e-5 \\
\bottomrule
\end{tabularx}
\end{table}

\begin{figure}[!tb]
\centering
\includegraphics[width=8.8cm]{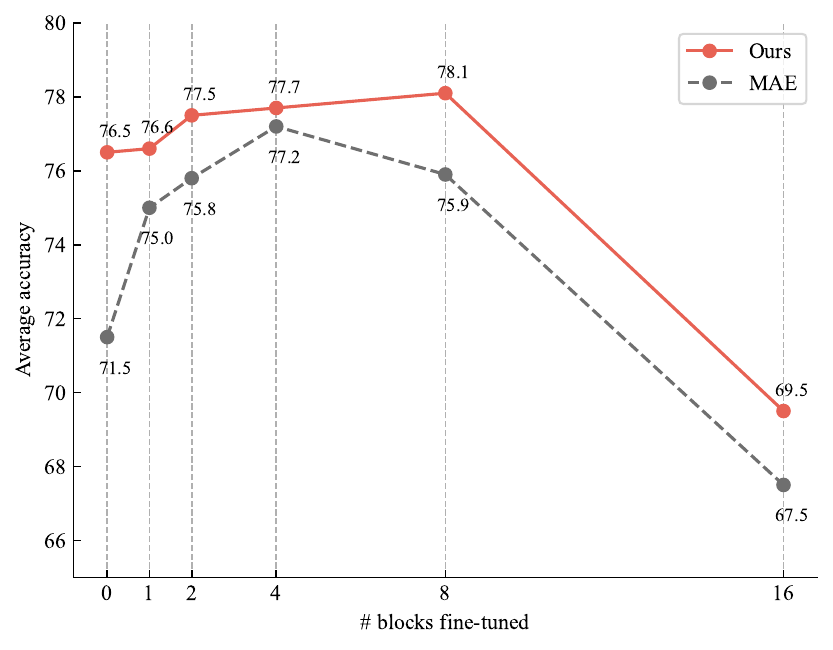}
\caption{Partial fine-tuning results of HiViT-B and SSL-ImageNet \& SAR. Tuning 0 blocks is linear probing. With the increasing number of fine-tuned transformer blocks, our method is consistently better than MAE and experiences overfitting later \textcolor{black}{under the 5-shot setting}.}
\label{fig_analysis_partial_finetuning}
\end{figure}

\begin{table*}[!tb]
\centering
\caption{Detailed classification results on MSTAR's SOCs and EOCs. ``-'' refers to unknown. \textbf{Bold} indicates the best result, \underline{underline} is the next best result.}
\label{table_detailed_results_1}
\renewcommand\arraystretch{1.2}
\resizebox{0.9\linewidth}{!}{%
\begin{tabular}{lc p{1.5cm}<{\centering} p{1.5cm}<{\centering}p{1.5cm}<{\centering}p{1.5cm}<{\centering}p{1.5cm}<{\centering}p{1.5cm}<{\centering}p{1.5cm}<{\centering}p{1.5cm}<{\centering}} 
\toprule
\multicolumn{10}{c}{SOCs: Standard operating conditions (10 way)} \\
\cmidrule(){1-10}
\multicolumn{1}{c}{Method} & Year & 1-shot & 2-shot & 3-shot & 4-shot & 5-shot & 10-shot & 20-shot & 25-shot \\
\cmidrule(lr){1-10}
DKTS-N~\cite{zhang2021domain} & 2021 & 49.3 & 58.5 & - & - & 72.3 & 84.6 & - & 96.2 \\
ConvT~\cite{Wang2022global} & 2022 & 42.6 & 54.4 & - & - & 75.2 & 88.6 & - & 96.5 \\ 
HDLM~\cite{Wang2022recognition} & 2022 & - & - & - & - & 72.4 
 & 88.2 & 95.2 & - \\
BIDFC~\cite{zhai2022weakly} & 2022 & \underline{80.7} & \underline{85.3} & \underline{87.3} & \underline{88.4} & \underline{90.3} & - & - & - \\
CRID~\cite{wang2023crucial} & 2023 & 48.3 & 51.0 & - & - & 73.3 & 87.4 & 96.9 & 97.1 \\
EUAPS~\cite{10138441} & 2023 & - & - & - & -  & 88.7 & \textbf{98.6} & \textbf{99.8} & - \\
PD~\cite{zhang2024optimal} & 2024 & 46.7 & 58.9 & 62.0 & 67.5 & 70.2 & 83.7 & - & - \\
\textbf{SARATR-X} & 2024 & \textbf{85.2} & \textbf{91.4} & \textbf{93.9} & \textbf{95.1}  & \textbf{95.9} & \underline{97.7} & \underline{98.1} & \textbf{98.5} \\
\cmidrule(){1-10}
\multicolumn{10}{c}{EOCs: Depression angle variations (4 way)} \\
\cmidrule(){1-10}
\multicolumn{1}{c}{Method} & Year & 1-shot & 2-shot & 3-shot & 4-shot & 5-shot & 10-shot & 20-shot & 25-shot \\
\cmidrule(lr){1-10}
DKTS-N~\cite{zhang2021domain} & 2021 & 61.9 & 63.9 & - & - & 67.4 & 71.1 & - & 78.9 \\
ConvT~\cite{Wang2022global} & 2022 & 59.6 & \underline{64.1} & - & - & 68.2 & 74.8 & \underline{79.1} \\
CRID~\cite{wang2023crucial} & 2023 & \underline{62.1} & 62.3 & - & - & \underline{74.5} & \underline{85.9} & - & \underline{87.0} \\
\textbf{SARATR-X} & 2024 & \textbf{93.4} & \textbf{97.3} & \textbf{98.5} & \textbf{98.0}  & \textbf{98.9} & \textbf{99.5} & \textbf{99.6} & \textbf{99.4} \\
\cmidrule(){1-10}
\multicolumn{10}{c}{EOCs: Target configuration variations (4-way)} \\
\cmidrule(){1-10}
\multicolumn{1}{c}{Method} & Year & 1-shot & 2-shot & 3-shot & 4-shot & 5-shot & 10-shot & 20-shot & 25-shot \\
\cmidrule(lr){1-10}
DKTS-N~\cite{zhang2021domain} & 2021 & 47.3 & 53.6 & - & - & 62.2 & 68.4 & - & 74.5 \\
ConvT~\cite{Wang2022global} & 2022 & 44.3 & 51.9 & - & - & 64.1 & \textbf{89.7} & - & \underline{91.0} \\
CRID~\cite{wang2023crucial} & 2023 & \underline{62.8} & \underline{65.7} & - & - & \underline{74.1} & 78.7 & - & 84.1 \\
\textbf{SARATR-X} & 2024 & \textbf{65.0} & \textbf{74.0} & \textbf{74.0} & \textbf{84.2}  & \textbf{78.3} & \underline{87.0} & \textbf{88.9} & \textbf{93.5} \\
\cmidrule(){1-10}
\multicolumn{10}{c}{EOCs: Target version configuration variations (4-way)} \\
\cmidrule(){1-10}
\multicolumn{1}{c}{Method} & Year & 1-shot & 2-shot & 3-shot & 4-shot & 5-shot & 10-shot & 20-shot & 25-shot \\
\cmidrule(lr){1-10}
DKTS-N~\cite{zhang2021domain} & 2021 & 48.9 & 55.1 & - & - & 65.6 & 70.2 & - & 77.0 \\
ConvT~\cite{Wang2022global} & 2022 & 42.3 & \underline{58.3} & - & - & \underline{68.1} & \underline{83.6} & - & \underline{92.0} \\
CRID~\cite{wang2023crucial} & 2023 & \underline{53.5} & 56.2 & - & - & 67.2 & 79.9 & - & 87.8 \\
\textbf{SARATR-X} & 2024 & \textbf{65.3} & \textbf{76.5} & \textbf{76.3} & \textbf{83.4}  & \textbf{82.8} & \textbf{85.0} & \textbf{92.1} & \textbf{97.0} \\
\bottomrule
\end{tabular}}
\end{table*}

We select three target classification datasets, including 25 fine-grained targets from vehicles, ships, aircraft, and others, to evaluate the comprehensive performance of SSL and the foundation model for SAR target recognition. The new SAR classification dataset is named SAR-Target in Table~\ref{table_class_dataset}.

\textbf{MSTAR}~\cite{MSTAR} is the most commonly used SAR vehicle datasetw. It has many experimental setting variants, while we refer to the ~\cite{10283916} to adopt the most commonly used ten-class classification settings, such as infantry vehicle, patrol car, personnel carrier, main battle tank, and truck.

\textbf{FUSAR-Ship}~\cite{hou2020fusar} contains 15 primary ship categories and many non-ship targets based on the Gaofen-3 satellite in scenes such as sea, land, coast, river, and island. Based on the experimental setting of~\cite{wang2022sar}, we have ten ocean target types, such as four fine-grained ships, bridges, ocean scenes, and false alarm slices. 

\textbf{SAR-ACD}~\cite{sun2022scan} contains five types of aircraft based on the Gaofen-3 satellite in three civil airports. Since the released dataset does not separate the training and test data, we randomly select partial samples as the training set and others as the test set. Fine-grained recognition of aircraft targets is a more challenging task due to the smooth surface of the aircraft, resulting in insignificant SAR image features.

\subsection{Hyperparameter settings}
\label{Hyperparameter settings}

\begin{table*}[!tb]
\centering
\caption{Detailed detection results. The metrics are mAP, $\rm{mAP_{50}}$ ($@$50), $\rm{mAP_{75}}$ ($@$75), $\rm{mAP_{small}}$ ($@$s), $\rm{mAP_{medium}}$ ($@$m), and $\rm{mAP_{large}}$ ($@$l). ``-'' refers to unknown. \textbf{Bold} indicates the best result, \underline{underline} is the next best result.}
\label{table_detailed_results_2}
\renewcommand\arraystretch{1.2}
\begin{tabular}{lc p{1.5cm}<{\centering} p{1.5cm}<{\centering}p{1.5cm}<{\centering}p{1.5cm}<{\centering}p{1.5cm}<{\centering}p{1.5cm}<{\centering}} 
\toprule
\multicolumn{8}{c}{SARDet-100K (Object detection)} \\
\cmidrule(){1-8}
\multicolumn{1}{c}{\multirow{1}{*}{Method}} & Year & mAP & $@$50 & $@$75 & $@$s & $@$m & $@$l \\
\cmidrule(lr){1-8} 
Deformable {DETR}~\cite{zhu2020deformable} & 2020 & 50.0 & 85.1 & 51.7 & 44.0 & 65.1 & 61.2 \\
Swin Transformer~\cite{liu2021swin} & 2021 & 53.8 & 87.8 & 59.0 & 49.1 & 64.6 & 60.0\\
VAN~\cite{guo2023visual} & 2022 & 53.5 & 86.8 & 58.0 & 47.3 & 65.5 & 60.6\\
ConvNext~\cite{liu2022convnet} & 2022 & 55.1 & 87.8 & 59.5 & 48.9 & 66.9 & 61.1\\
MSFA~\cite{li2024sardet100k} & 2024 & \underline{56.4} & \underline{88.2} & \underline{61.5} & \underline{50.5} & \underline{66.5} & \underline{62.5}\\
\textbf{SARATR-X} & 2024 & \textbf{57.2} {\scriptsize (+0.8)} & \textbf{88.5} {\scriptsize (+0.3)} & \textbf{62.6} {\scriptsize (+1.1)} & \textbf{51.2} {\scriptsize (+0.7)} & \textbf{69.6} {\scriptsize (+3.1)} & \textbf{64.7} {\scriptsize (+2.2)}\\
\cmidrule(){1-8}
\multicolumn{8}{c}{OGSOD\tnote{*} (Object detection)} \\
\cmidrule(){1-8}
\multicolumn{1}{c}{\multirow{1}{*}{Method}} & Year & mAP & $@$50 & $@$75 & $@$s & $@$m & $@$l \\
\cmidrule(lr){1-8} 
Generalized Focal~\cite{he2019bounding} & 2019 & 41.8 & 67.6 & - & - & - & - \\
Sparse R-CNN~\cite{sun2021sparse} & 2021 & 38.7 & 65.6 & - & - & - & -\\
Object Box~\cite{zand2022objectbox} & 2022 & 40.1 & 76.6 & - & - & - & -\\
YOLOv7~\cite{wang2023yolov7} & 2022 & \underline{45.1} & \underline{79.2} & - & - & - & - \\
\textbf{SARATR-X} & 2024 & \textbf{51.4} {\scriptsize (+6.3)} & \textbf{85.2} {\scriptsize (+6.0)} & \textbf{51.7} & \textbf{46.8} & \textbf{75.3} & \textbf{77.1}\\
\cmidrule(){1-8}
\multicolumn{8}{c}{SSDD (Ship detection)} \\
\cmidrule(){1-8}
\multicolumn{1}{c}{\multirow{1}{*}{Method}} & Year & AP & $@$50 & $@$75 & $@$s & $@$m & $@$l \\
\cmidrule(lr){1-8} 
FBR-Net~\cite{fu2021anchor} & 2021 & - & 94.1 & 59.1 & - & - & -\\
CenterNet++~\cite{guo2021centernet++} & 2021 & - & 95.1 & - & - & - & -\\
CRTransSar~\cite{xia2022crtranssar} & 2022 & - & 97.0 & \underline{76.2} & - & - & -\\
YOLO-Lite~\cite{ren2023yolo} & 2023 & - & 94.4 & - & - & - & -\\
FEPS-Net~\cite{bai2023feature} & 2023 & 59.9 & 96.0 & 67.5 & \underline{55.1} & \textbf{68.2} & \textbf{70.6}\\
$\rm{CS}^n$Net~\cite{Chen2023CSnNet} & 2023 & \underline{64.9} & \underline{97.1} & - & - & - & -\\
\textbf{SARATR-X} & 2024 & \textbf{67.4} {\scriptsize (+2.5)} & \textbf{97.3} {\scriptsize (+0.2)} & \textbf{83.1} {\scriptsize (+6.9)} & \textbf{68.1} {\scriptsize (+13.0)} & \underline{66.1} {\scriptsize (-2.1)} & \underline{60.5} {\scriptsize (-10.1)}\\
\cmidrule(){1-8}
\multicolumn{8}{c}{SAR-Aircraft (Aircraft detection)} \\
\cmidrule(){1-8}
\multicolumn{1}{c}{\multirow{1}{*}{Method}} & Year & AP & $@$50 & $@$75 & $@$s & $@$m & $@$l \\
\cmidrule(lr){1-8} 
Cascade R-CNN~\cite{cai2018cascade} & 2018 & - & 75.7 & 58.9 & - & - & - \\
RepPoints~\cite{yang2019reppoints} & 2019 & - & 72.6 & 53.3 & - & - & - \\
SKG-Net~\cite{fu2021scattering} & 2021 & - & 70.7 & 46.4 & - & - & - \\
SA-Net~\cite{wang2023sar} & 2023 & - & \underline{77.7} & \underline{62.8} & - & - & - \\
\textbf{SARATR-X} & 2024 & \textbf{58.7} & \textbf{86.1} {\scriptsize (+5.7)} & \textbf{64.7} {\scriptsize (+3.3)} & \textbf{20.0} & \textbf{57.2} & \textbf{49.7} \\
\bottomrule
\end{tabular}
\end{table*}

Here are the detailed settings of our pre-training and downstream tasks as shown in Table~\ref{table: hyperparameter settings pre-training} and~\ref{table: hyperparameter settings classification}.

\textbf{Pre-training -} 
Our default pre-training setting is in Table~\ref{table: hyperparameter settings pre-training}, and other hyperparameters of each method use the default settings from their papers and codes. Our pre-training is applied on 8 NVIDIA RTX3090 GPUs with 200 epochs and 800 batch sizes. Compared to the training settings of MAE, we add ColorJitter ($\rm{contrast}=0.5$) to increase data richness. Moreover, we modify the batch size and epoch according to 8 GPUs. It is worth noting that although MAE uses the normalized pixel value to enhance the feature representation in the visible spectral images, we find that the normalized pixel value cannot be used due to the SAR image noise and prevents the training loss from decreasing properly. 

\textbf{Classification setting -} 
All models use the same training settings in downstream classification tasks. Table~\ref{table: hyperparameter settings classification} gives the default setting. Our few-shot learning setting is based on the Dassl toolbox~\cite{zhou2022domain,zhou2021domain} and is averaged over 10 random experiments. Since we focus on the small sample case of the downstream classification task, we use the linear probing method in MAE to finetune models and avoid overfitting.

\textbf{Partial fine-tuning -} Fig.~\ref{fig_analysis_partial_finetuning} shows why we chose linear probing for few-shot evaluation, as HiViT overfits when many blocks are fine-tuned. Experimental results show that our method consistently obtains better representations than MAE, and overfitting occurs later.

\section{Implementation Details for Section 5 and the Leveraging of SARATR-X for Recognition Tasks}
\label{downstream setting}

\subsection{Dataset description}
We choose MSTAR, the most commonly used dataset in SAR target classification. SOCs are in similar image conditions, and the training set's depression angle under SOC is 17°, while the test set is 15°. Ten categories of targets include BMP2, BRDM2, BTR60, BTR70, T62, T72, 2S1, D7, ZIL131, and ZSU234 as shown in Table~\ref{table_class_dataset}. EOCs settings are the imaging condition variations to test robustness. Existing methods have been saturated with different experimental settings of MSTAR~\cite{10283916}, but few samples and depression angle variations remain challenging. 

We use SARDet-100K and OGSOD datasets, which have many test samples and categories, to evaluate the detection performance fully. The OGSOD comparison results are derived from the original article's~\cite {wang2023category} single-modal approach using only SAR images. SSDD and SAR-Aircraft are ship and aircraft categories. 

\subsection{Hyperparameter settings}
Based on our scaling experiment, we use HiViT-B with 600 epochs pre-trained on SSL-ImageNet \& SAR as the foundation model for classification and detection tasks.

\textbf{Classification setting} is follow Table~\ref{table: hyperparameter settings classification}. The only difference is that we use partial fine-tuning for better performance, and the last 6 blocks are used to the fine-tuned.

\textbf{Detection setting} follows the default setting in HiViT, and we adjust the learning rate to 5e-4. We used the same settings for each dataset fine-tuning, and our GitHub configuration is based on the mmdetection~\cite{chen2019mmdetection} framework for details.

\subsection{Detailed results}
We provide detailed classification and detection results in Table~\ref{table_detailed_results_1} and~\ref{table_detailed_results_2}. Although the proposed method outperforms existing methods in mAP, there is still scope for improvement in some refined metrics.

\bibliographystyle{IEEEtran}
\bibliography{ref}

\end{document}